\documentclass[11pt]{article}

\usepackage[preprint]{acl}

\usepackage{times}
\usepackage{latexsym}

\usepackage[T1]{fontenc}

\usepackage[utf8]{inputenc}

\usepackage{microtype}

\usepackage{inconsolata}

\usepackage{graphicx}

\usepackage{url}
\usepackage{dirtytalk}
\usepackage{booktabs}
\usepackage{multirow}
\usepackage{amsmath}

\usepackage{tikz}
\usetikzlibrary{arrows.meta,positioning,calc,shapes.misc}
\usepackage{xcolor}

\definecolor{prismaBlue}{RGB}{31,78,121}
\definecolor{prismaGray}{RGB}{89,89,89}
\definecolor{prismaGreen}{RGB}{56,118,29}

\definecolor{DarkBlue}{RGB}{0,82,147}
\definecolor{LightBlue}{RGB}{100,160,200}
\definecolor{LighterBlue}{RGB}{152,198,234}

\title{PrivBench: A Holistic and Modular Benchmarking Platform for Evaluating Text-to-Text Privatization}

\author{Stephen Meisenbacher, Andreea-Elena Bodea, Ahmet Bilal Akın,\\ \textbf{Alexandra Klymenko, Jana Diesner, \and Florian Matthes} \\
Technical University of Munich\\
School of Computation, Information and Technology \\
Department of Computer Science\\
Garching, Germany\\
\normalsize\texttt{\{stephen.meisenbacher,andreea.bodea,bilal.akn,alexandra.klymenko,matthes\}@tum.de} \\
}

\author{
 \textbf{Stephen Meisenbacher\textsuperscript{1,2,3,4}}{\normalfont ,}
 \textbf{Andreea-Elena Bodea\textsuperscript{2}}{\normalfont ,}
 \textbf{Ahmet Bilal Akın\textsuperscript{1}}{\normalfont ,}
 \\
 \textbf{Alexandra Klymenko\textsuperscript{1}},
 \textbf{Jana Diesner\textsuperscript{1,2,3,4}}, \and 
 \textbf{Florian Matthes\textsuperscript{1}}
\\
 \textsuperscript{1}Technical University of Munich, School of Computation, Information and Technology\\
 \textsuperscript{2}Technical University of Munich, School of Social Sciences and Technology\\
 \textsuperscript{3}Munich Center for Machine Learning, Munich, Germany\\
 \textsuperscript{4}Munich Data Science Institute, Munich, Germany\\
 \small{
   \textbf{Correspondence:} \href{mailto:stephen.meisenbacher@tum.de}{stephen.meisenbacher@tum.de}
 }
}

\begin{document}
\maketitle
\begin{abstract}
Natural Language Processing methods have enabled novel solutions and advances in the field of privacy, particularly in the sub-domain of \textit{text-to-text privatization}, where the goal is to transform a sensitive input text into a privatized output by ideally masking (in)directly identifiable or otherwise private information. The evaluation of text-to-text privatization, however, is not straightforward, and the extant literature has utilized a myriad of techniques and metrics to quantify the privacy-preserving capabilities of privatization methods. Seeking to unify the evaluation of text-to-text privatization, we introduce PrivBench, a holistic and modular benchmarking platform for researchers and practitioners working on text privatization. PrivBench is holistic in that it evaluates privatization on a series of defined desiderata, which are structured into \textit{modules}. PrivBench is not only modular but also extensible, allowing for future updates and benchmark versions. PrivBench is user-centered and promotes competition via real-time evaluation and a live public leaderboard. The platform is free to use and openly accessible at \url{https://privbench.com/}.
\end{abstract}

\section{Introduction}
The field of privacy-preserving Natural Language Processing (PPNLP) has grown in recent years \cite{habernal-etal-2023-privacy, sousa2023keep}, in line with the increase in NLP and LLM applications. Under the umbrella of PPNLP fall a variety of text privatization methods, including classic anonymization, Named Entity Recognition-based techniques, and modern methods leveraging LLMs as text privatizers \cite{11247969}. Of particular interest are \textit{text-to-text} privatization methods, where an input text is transformed, rewritten, or changed in a manner that preserves privacy, while also ideally maintaining the original usefulness and meaning of the text \cite{mattern-etal-2022-limits}.

While the design of effective text-to-text privatization methods is challenging in its own right, the \textit{evaluation} of these methods presents an additional hurdle \cite{ren-etal-2025-measure}: beyond the general difficulty with accurately incorporating societal notions of \say{privacy} into evaluation procedures, the particular field of text privatization must not only measure privacy protection, but also balance privacy with utility preservation, often referred to as the \textit{privacy-utility trade-off} \cite{loiseau-etal-2025-tau}. In this confluence of privacy and utility, numerous evaluation approaches have been designed for text privatization, with little to no uniformity \cite{ren-etal-2025-measure,loiseau-etal-2025-tau}.

This lack of standardization is coupled with a lack of accessible mechanisms to comparatively analyze the benefits and limitations of text-to-text privatization methods across a wide variety of disciplines, such as anonymization \cite{pilan-etal-2022-text}, Differential Privacy \cite{klymenko-etal-2022-differential}, or synthetic data \cite{yue-etal-2023-synthetic}. This differs from other sub-fields in NLP that have long-standing and competitive benchmark frameworks, such as GLUE and MMLU for Natural Language Understanding \cite{wang-etal-2018-glue,hendrycks2021measuringmassivemultitasklanguage}, or BIG-Bench \cite{srivastava2023beyond} and HELM \cite{liangholistic} for more comprehensive Language Model benchmarking.

To address these shortcomings, we introduce PrivBench, a fully operational benchmarking platform that centers on the holistic evaluation of text-to-text privatization methods across a range of privacy and utility evaluation modules. 
The platform hosts and runs these benchmarks in real time upon user submission.
PrivBench features the ability to create and manage user accounts, and each user can choose to feature their submissions on a public leaderboard, allowing for competition and comparability. Because the platform is modular, it is also extensible, allowing for the addition, editing, or removal of evaluation modules in future updates, tracked by a versioning system. Thus, the mission of PrivBench is not only to provide a unified benchmarking resource, but also to foster community engagement and growth.
Our contributions are:
\begin{enumerate}
    \itemsep -0.3em
    \item We extend previous work with a comprehensive survey of evaluation techniques and metrics for text-to-text privatization.
    \item We leverage these theoretical foundations to design and implement a fully operational and deployed benchmarking platform: PrivBench.
    \item We make the platform openly accessible, free, and available for immediate use, found at \url{https://privbench.com/}.
    \item  We test our holistic benchmark in a comparative case study of 14 privatization methods.
\end{enumerate}

\section{Related Work}
Although the core of prior work in text privatization focuses on the evaluation of the proposed methods, some recent papers either systematize evaluation in general, or propose new benchmarks for text privatization.
In a recent survey, \citet{ren-etal-2025-measure} review 47 related publications to investigate how privacy in text is evaluated, and categorize privacy metrics into six categories. Their survey sheds light on the multi-faceted nature of text privatization evaluation. Other surveys also highlight the importance of evaluation in PPNLP \cite{klymenko-etal-2022-differential,igamberdiev-etal-2022-dp,sousa2023keep}.

Benchmark frameworks have been published in other PPNLP sub-domains. Focusing on privacy in language models, PrivLM-Bench \cite{li-etal-2024-privlm}, PrivacyLens \cite{shao2024privacylens}, and PrivAuditor \cite{zhu2024privauditor} have been proposed. However, these benchmarks do not address text-to-text privatization. Other papers propose benchmarks for specific frameworks like Differential Privacy \cite{igamberdiev-etal-2022-dp,meisenbacher-etal-2024-comparative,sun2025synbenchbenchmarkdifferentiallyprivate}, which lack generalizability to the wider spectrum of text-to-text privatization.

More recently, \citet{loiseau-etal-2025-tau} proposed TauEval, a framework for the unified evaluation of text-to-text privatization approaches. Their framework evaluates privatization methods across eight different tasks, which focus on different aspects of \textit{utility} given privatized texts. As with \citet{ren-etal-2025-measure}, the importance of multi-dimensional privacy evaluation is supported. However, \citet{loiseau-etal-2025-tau} emphasize utility-based evaluation over privacy metrics, and the choice of metrics is not grounded in a review of literature. Furthermore, TauEval is released as a Python software package without a public-facing platform, as is present in benchmarks for many other NLP fields.

We build on these prior contributions by proposing a comprehensive and holistic text-to-text privatization benchmark that is (1) modular to reflect the many dimensions of privacy evaluation, (2) publicly accessible, no-code, and interactive to promote competition and to connect researchers, and (3) extensible to allow for improvements and version updates as text privacy evaluation evolves.

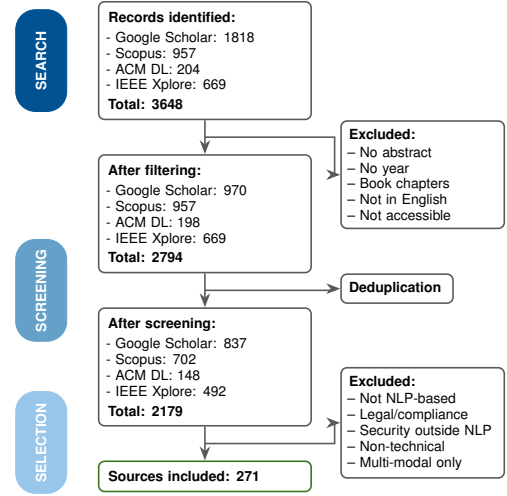
\begin{figure}[t]
\centering
\resizebox{0.84\columnwidth}{!}{%
\begin{tikzpicture}[
  font=\sffamily\scriptsize,
  >={Stealth[length=2.5mm]},
  node distance=6mm,
  main/.style={
    draw=prismaGray,
    thick,
    fill=white,
    rounded corners=3pt,
    inner sep=5pt,
    align=left,
    text width=34mm
  },
  finalbox/.style={
    draw=prismaGreen,
    thick,
    fill=white,
    rounded corners=3pt,
    inner sep=5pt,
    align=left,
    text width=34mm
  },
  callout/.style={
    draw=prismaGray,
    thick,
    fill=white,
    rounded corners=3pt,
    inner sep=4pt,
    align=left,
    text width=26mm,
    font=\sffamily\scriptsize
  },
  smallcallout/.style={
    draw=prismaGray,
    thick,
    fill=white,
    rounded corners=3pt,
    inner sep=4pt,
    align=center,
    font=\sffamily\scriptsize
  },
  pill/.style={
    draw=LightBlue,
    fill=LightBlue,
    text=white,
    rounded corners=10pt,
    minimum height=9mm,
    minimum width=18mm,
    rotate=90,
    align=center,
    font=\sffamily\scriptsize\bfseries
  },
  arrow/.style={
    ->,
    thick,
    prismaGray
  }
]
\node[main] (b1) {%
  \textbf{Records identified:}\\[2pt]
  - Google Scholar: 1818\\
  - Scopus: 957\\
  - ACM DL: 204\\
  - IEEE Xplore: 669\\[2pt]
  \textbf{Total: 3648}
};
\node[main, below=of b1] (b2) {%
  \textbf{After filtering:}\\[2pt]
  - Google Scholar: 970\\
  - Scopus: 957\\
  - ACM DL: 198\\
  - IEEE Xplore: 669\\[2pt]
  \textbf{Total: 2794}
};
\node[main, below=of b2] (b3) {%
  \textbf{After screening:}\\[2pt]
  - Google Scholar: 837\\
  - Scopus: 702\\
  - ACM DL: 148\\
  - IEEE Xplore: 492\\[2pt]
  \textbf{Total: 2179}
};
\node[finalbox, below=of b3] (b4) {%
  \textbf{Sources included: 271}\\[2pt]
};
\node[pill, fill=DarkBlue, draw=DarkBlue] (p1) at ($(b1.west)+(-10mm,0)$) {SEARCH};
\node[pill, fill=LightBlue, draw=LightBlue] (p2) at ($(b2.south west)!0.5!(b3.north west)+(-10mm,0)$) {SCREENING};
\node[pill, fill=LighterBlue, draw=LighterBlue] (p3) at ($(b4.west)+(-10mm,0.6)$) {SELECTION};
\draw[arrow] (b1.south) -- (b2.north);
\draw[arrow] (b2.south) -- (b3.north);
\draw[arrow] (b3.south) -- (b4.north);
\node[callout, right=5mm of b1.south east, anchor=north west] (cFilter) {%
  \textbf{Excluded:}\\[1pt]
  -- No abstract\\
  -- No year\\
  -- Book chapters \\
  -- Not in English \\
  -- Not accessible
};
\node[smallcallout, right=5mm of b2.south east, anchor=north west] (cDedup) {%
  \textbf{Deduplication}
};
\node[callout, right=5mm of b3.east, anchor=north west] (cExcl) {%
  \textbf{Excluded:}\\[1pt]
  -- Not NLP-based\\
  -- Legal/compliance\\
  -- Security outside NLP\\
  -- Non-technical\\
  -- Multi-modal only
};
\coordinate (mid12) at ($(b1.south)!0.5!(b2.north)$);
\coordinate (mid23) at ($(b2.south)!0.5!(b3.north)$);
\coordinate (mid34) at ($(b3.south)!0.5!(b4.north)$);
\draw[arrow] (mid12) -| ($(cFilter.west)+(-1mm,0)$) -- (cFilter.west);
\draw[arrow] (mid23) -| ($(cDedup.west)+(-1mm,0)$) -- (cDedup.west);
\draw[arrow] (mid34) -| ($(cExcl.west)+(-1mm,0)$) -- (cExcl.west);
\end{tikzpicture}%
}
\caption{Systematic literature review selection process.}
\label{fig:slr-process}
\end{figure}

\begin{table*}[t]
\centering
\small
\resizebox{\linewidth}{!}{
\begin{tabular}{p{0.08\textwidth}|p{0.115\textwidth}|p{0.9\textwidth}|p{0.03\textwidth}|p{0.35\textwidth}|p{0.09\textwidth}}
\hline
\textbf{Category} & \textbf{Module} & \textbf{Description} & \textbf{\#} & \textbf{Representative Papers} & \textbf{Dataset(s)} \\
\hline
\multirow{15}{*}{Privacy} 
& Attribute\newline Inference (ATT) & Measures resistance to authorship inference attacks. A pretrained dataset-specific DeBERTa authorship identification classifier\footnotemark[1] is run on both the original and privatized texts, and the micro-F1 scores are compared. The score reflects how much the classifier's accuracy degrades after privatization; a higher score means better privacy protection. & 31 & \cite{macko2024authorship, meisenbacher2024spend} & Reddit, Yelp \\
\cline{2-6}
& Masked Token Inference (MTI) & Measures resistance to token-level inference attacks. Each token in the privatized text is masked one at a time, and a masked language model attempts to predict the masked token. The predicted tokens are then compared against the corresponding original text tokens. & 5 & \cite{tong2025inferdpt,meisenbacher2024spend,chen-etal-2023-customized,yue-etal-2021-differential} & WikiText, TAB\\
\cline{2-6}
& Nearest\newline Neighbor\newline Privacy (NNP) & Measures resistance to record-linkage attacks by testing whether original texts can be matched back to their privatized counterparts through semantic similarity. Texts are embedded using a sentence transformer, and for each original text, its nearest neighbor rank among the privatized texts is computed. A higher score indicates that linking original to privatized texts is difficult. & 3 & \cite{meisenbacher2024spend,carvalho2023tem,xu-etal-2021-utilitarian} & WikiText, Yelp\\
\cline{2-6}
& Private Entity Masking (PEM) & Measures the effectiveness of named entity removal. A spaCy named entity recognition model is applied to the original texts to identify all named entities. Each detected entity is then checked for presence in the corresponding privatized text using case-insensitive matching. The score represents the proportion of original entities that were successfully removed or obfuscated. & 11 & \cite{vinod2025invisibleink, Mancera_Morales_2026, vats2024recovering, smith2024identifying, qu2021natural} & WikiText, TAB \\
\cline{2-6}
& Length\newline Robustness (LRO) & Measures the consistency of a module across different text lengths. Texts are grouped into bins by word count, and the module is evaluated separately on each bin. The final score combines the average performance across bins with a consistency penalty that penalizes large variation. & - & Meta-metric: builds on top of another module (in v1.0.0: Private Entity Masking). & Yelp \\
\hline\hline
\multirow{15}{*}{Utility} 
& Text\newline Coherence (COH) & Measures the naturalness and fluency of privatized texts using perplexity of a GPT-2 model. Perplexity is computed for both the original and privatized texts, and the score reflects the ratio between them. & 37 & \cite{awon2025clusant, flemings2024differentially_private_next_token, vats2024recovering, meisenbacher2024thinking, li2023beyond} & IMDb, WikiText \\
\cline{2-6}
& MAUVE (MAU) & Measures the distributional similarity between the original and privatized text collections as a whole. Rather than comparing individual text pairs, MAUVE evaluates whether the overall distribution of privatized texts resembles the distribution of originals in the embedding space. & 12  & \cite{tong2025inferdpt, vinod2025invisibleink, carranza2024synthetic, xie2024differentially, uchendu2023attribution} & WikiText, Yelp\\
\cline{2-6}
& Similarity (SIM) & Measures how well the privatized texts preserve the meaning and content of the originals. Two complementary metrics are combined: BLEU score captures lexical overlap at the n-gram level, while cosine similarity between sentence embeddings captures semantic preservation. The final score is the average of both metrics. & 49 & \cite{tong2025inferdpt, higashi2025differential, awon2025clusant, huang2024private, meisenbacher2024dp, yu2021differentially} & PubMedQA, Yelp\\
\cline{2-6}
& Task\newline Utility (TUT) & Measures the degree to which downstream task performance can be preserved after privatization. A fine-tuned model for sentiment analysis\footnotemark[2] (IMDb) or question answering\footnotemark[3] is run on both the original and privatized texts, and the micro-F1 scores are compared. The score reflects how much the classifier's accuracy degrades after privatization. & 147\footnotemark[4] & The large majority of works perform some manner of downstream task utility. & PubMedQA, IMDb \\
\cline{2-6}
& Length\newline Variation (LVA) & Measures whether privatized texts maintain appropriate length compared to the originals. The module computes the ratio of privatized to original text length for each pair and evaluates both the mean ratio and its variance. A mean ratio close to one indicates that privatized texts are neither dramatically shorter nor longer than the originals. & - &  Meta-metric: newly proposed, motivated by the discussions of \citet{mattern-etal-2022-limits}, \citet{meisenbacher2024dp}, and others. & WikiText, Yelp \\
\bottomrule
\end{tabular}
}
\caption{Overview of the 10 evaluation modules of PrivBench v1.0.0. Modules are balanced between \textit{privacy} and \textit{utility} metrics. \textit{\#} indicates the number of papers in which this metric was utilized, according to our literature review.}
\label{tab:modules}
\end{table*}

\section{Evaluating Text Privatization}
\paragraph{Literature review.}
To ground our platform in a holistic evaluation of text-to-text privatization, we conducted a systematic literature review of 271 papers. From this, we aggregate evaluation strategies to inform the modular approach of PrivBench.

We conducted our literature review according to the systematic approach outlined by \citet{10.5555/2994449}. The first step was the design of a search string to probe relevant academic literature databases. This was done collaboratively among our research team, where the focus was to include key terms that indicate NLP \textit{methods} \underline{and} privacy. The general focus was scoped to ensure a comprehensive selection of publications. The final search string was:

\begin{center}
    \scriptsize
    \textit{("Natural language processing" OR "NLP" OR "Language models" OR "Language" OR "Text" OR "Document" OR "Information extraction" OR "Named entity recognition" OR "Part-of-speech tagging" OR "Sentiment analysis" OR "Machine translation" OR "Information retrieval") AND ("Privacy" OR "Private" OR "Privatization" OR "Data privacy" OR "Privacy-preserving" OR "Privacy-enhancing technologies" OR "PETs" OR "Privacy concerns" OR "Data protection" OR "Anonymization" OR "Rewriting" OR "Sanitization" OR "Obfuscation" OR "Scrubbing")}
\end{center}

Using this string, we queried four prominent academic databases for technical papers: \textit{Google Scholar}, \textit{Scopus}, \textit{ACM Digital Library}, and \textit{IEEEXplore}. Note that the ACL Anthology is indexed by Google Scholar. We queried for title only, in order to maintain a manageable number of results. The final search was performed in June 2026.

\footnotetext[1]{\scriptsize\url{https://huggingface.co/sjmeis/yelp_authorship_small} and \url{https://huggingface.co/sjmeis/reddit-mental-health_authorship/}}
\footnotetext[2]{\scriptsize\url{https://huggingface.co/sjmeis/imdb_sentiment_small}}
\footnotetext[3]{\scriptsize\url{https://huggingface.co/sjmeis/pubmedqa_base/}}
\footnotetext[4]{\scriptsize Based on the \textit{Accuracy} metric alone.}

From the four databases, the results (3648 in total) were filtered for sources without a year or abstract, published as a book, not written in English, not publicly accessible, or otherwise clearly lacking quality. The resulting set (2794) were combined and deduplicated based on title. This resulted in a filtered set of 2179 papers. These papers were all screened by three researchers, who read the title and abstract of each paper and applied inclusion/exclusion criteria to vote on candidacy for inclusion. The primary inclusion criteria was sources researching text-to-text privatization, whereas exclusion criteria are outlined in Figure \ref{fig:slr-process}.

Papers receiving a unanimous three votes were automatically included. Any paper receiving one or two votes was reviewed again by the primary researcher, who made an executive decision based on team discussion. This process resulted in a final set of 271 candidates for full-text reading.

All 271 papers were read by one researcher (as the qualitative coding required in-depth domain knowledge) \cite{bernard2016analyzing}, who screened specifically for evaluation approaches undertaken in the publication. These approaches were tabulated, enumerated, and later aggregated. The datasets used for evaluation were also noted. 40 papers were double-coded by the primary researcher to ensure agreement. All extracted evaluation methods can be found in Table \ref{tab:metrics_counts} of the Appendix; there are 35 in total, and they are divided into six categories. We note that only evaluation approaches appearing in \textgreater2 papers were included, in order to manage a large number of singleton approaches.

From this, we hand-selected 10 evaluation approaches to be operationalized in the first version of PrivBench, justified and briefly introduced next.

\paragraph{A holistic set of text-to-text privatization evaluation techniques.}
The 10 distinct approaches represent a diverse selection across privacy and utility metrics appropriate for text-to-text privatization. These were selected in favor of approaches that have been more widely adopted (and thus validated by researcher adoption), and to balance the evaluation of privacy protection \textit{and} utility preservation.

From the selected metrics, we sought to find a minimal set of datasets that have been used for the respective metrics in the literature. This resulted in six total datasets. All 10 approaches are introduced in Table \ref{tab:modules} with representative sources and corresponding datasets. These form the foundation of the evaluation \textit{modules} of PrivBench.

\section{System Design}
PrivBench is implemented as a full-stack web application, utilizing a JavaScript (React) frontend, a Python (Flask) backend, a PostgreSQL database, and a Celery task management system with Redis.

\paragraph{System architecture.}
The web application is set up as a traditional full-stack application with a frontend, backend, and persistent database. The application is running on a local server with a 32-core CPU, 32 GB of RAM, and a single Nvidia RTX 5060 TI 16GB GPU, as well as 2TB of storage. Future plans include to migrate the system to a setup that can facilitate more concurrent requests.

The web application is dockerized for easy management and deployment. Each module is spawned as its own separate Docker container, allowing for task queuing and parallel benchmarking. Celery is used as a distributed task queue, where a user submission is distributed to all modules for evaluation. Redis is utilized as a message broker between the application and Celery workers.

PrivBench is made publicly accessible via nginx, which is mapped to the custom \url{https://privbench.com} domain. The site's SSL certificate is checked or renewed daily via certbot.

\begin{figure*}[t]
    \centering
    \includegraphics[width=0.99\linewidth]{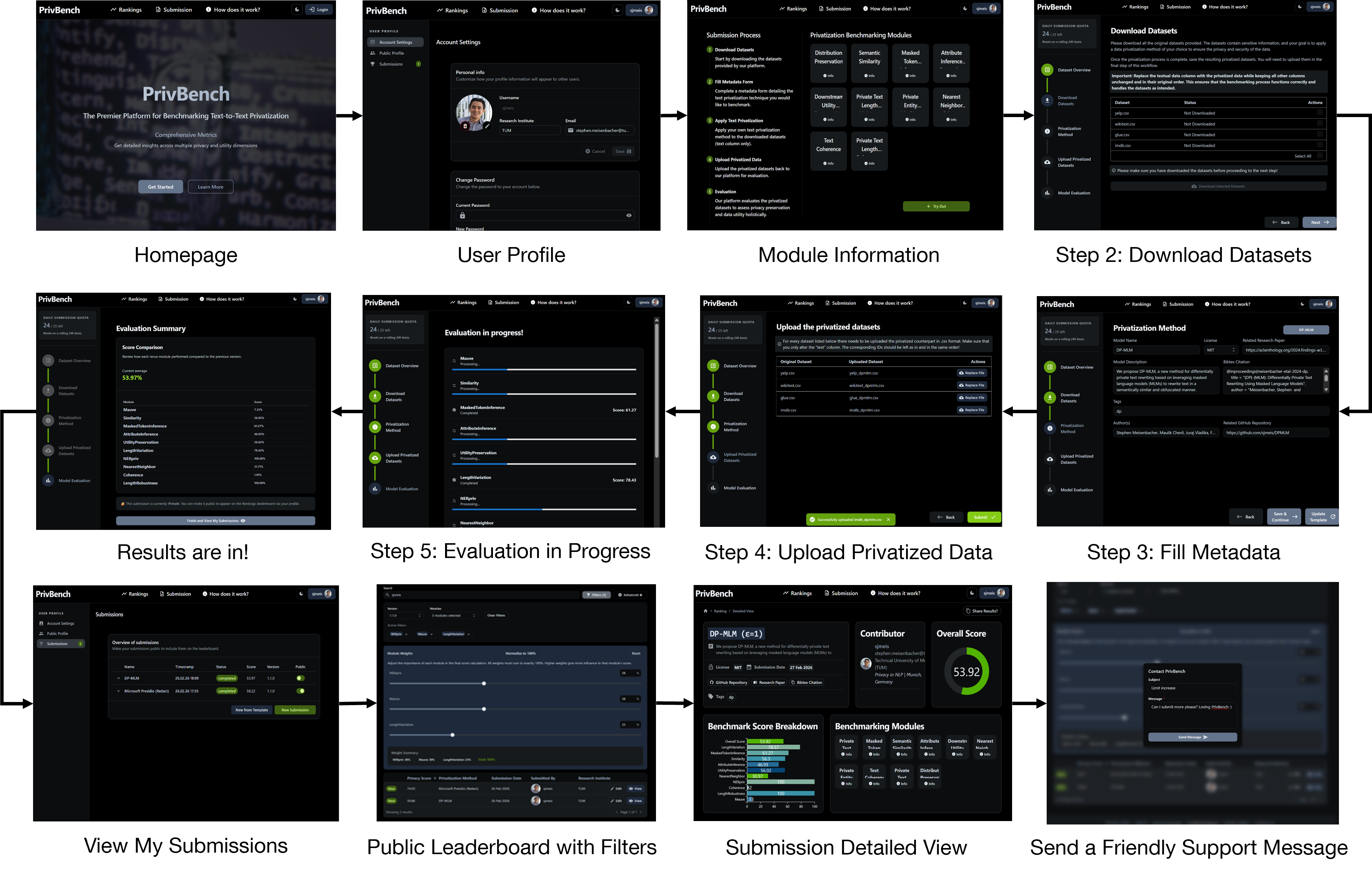}
    \caption{The \textit{researcher} user journey.}
    \label{fig:user}
\end{figure*}

\begin{figure*}[!ht]
    \centering
    \includegraphics[width=0.99\linewidth]{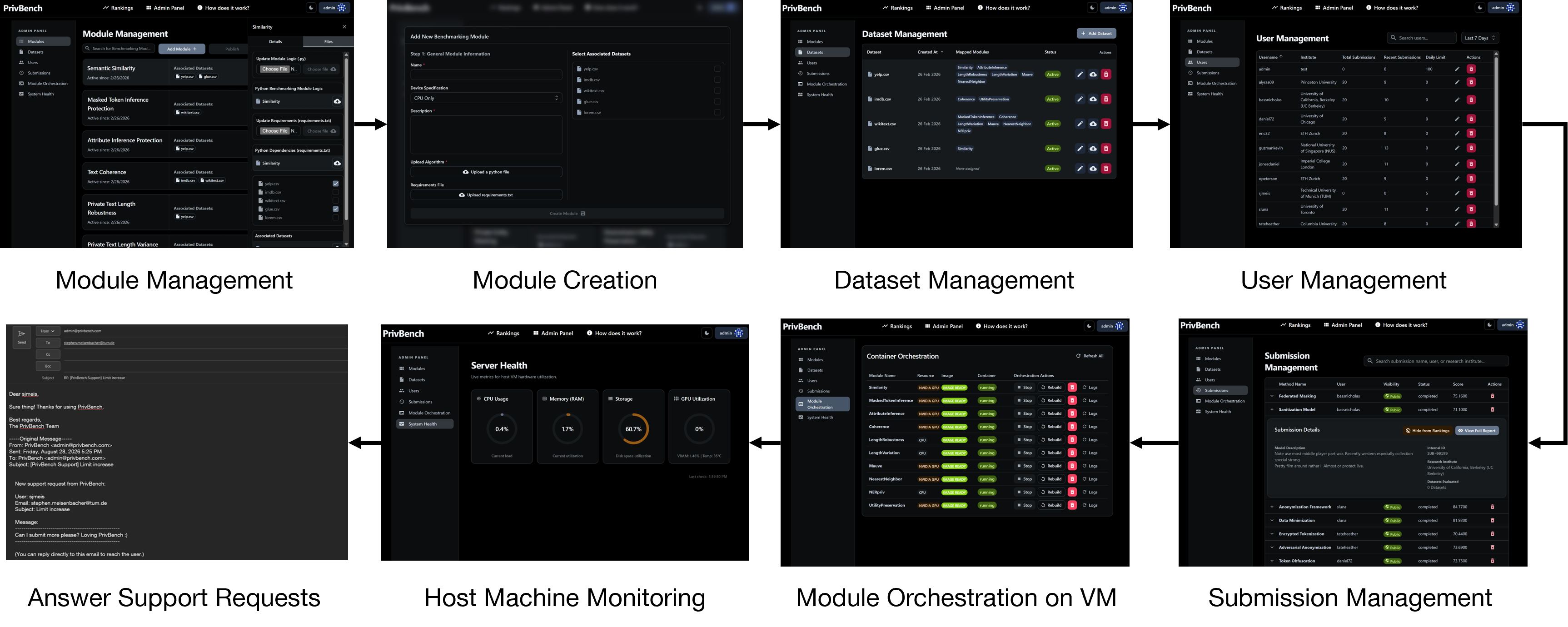}
    \caption{The \textit{admin} user journey.}
    \label{fig:admin}
\end{figure*}

\paragraph{User personas and user journey.}
For PrivBench, we envision two primary user personas: the researcher/practitioner (henceforth, researcher) and the administrator (admin). Their user journeys are narrated in the following, and these descriptions are mirrored by the snapshots of the PrivBench experience, provided in Figures \ref{fig:user} and \ref{fig:admin}.

\textbf{Researcher.} 
The (privacy) researcher is the primary target audience of PrivBench, where the goal is to enable a standardized evaluation of text privatization methods in development, which in turn enables comparability to existing methods.

The researcher starts the user journey by registering an account on the site, which must be verified via a unique code sent to the provided email address. Password resets are also possible. Once the user is verified, the account profile can be accessed, where a profile picture can be uploaded, and a short bio can be provided. In the account area, all user submissions (discussed next) can also be viewed.

The researcher sees an overview page of the current modules by clicking on \textit{How does it work?} on the navigation bar. Then, to start a benchmarking submission, the \textit{Submission} tab can be clicked. This page outlines a clear five-step submission process: (1) informational overview, (2) dataset download (i.e., the original text datasets to be privatized), (3) metadata input for a text-to-text privatization method (including from a saved template), (4) privatized dataset upload, and (5) evaluation progress overview (i.e., after submission). The user is informed if placed in the queue for any module (implying other user(s) are currently ahead in line).

Following the successful completion of a benchmarking run, an overview of the results is presented. The researcher is prompted to head to the account area to view the result, where there exists the option to make the result public, thereby placing it on the public leaderboard. After this, the researcher can then navigate to the \textit{Rankings} tab, which houses the leaderboard. There, users can search and filter the table, and as an advanced feature, the scores can be modified via user-defined weights for each of the modules. Each public result can also be explored in a \say{detailed view}, which gives an in-depth overview of the submission and scores.

Finally, navigating to the footer, the researcher can view PrivBench's version history, which outlines all published changes. There is also the ability to contact the admin team via a simple form, which sends a message to the PrivBench email inbox.

\textbf{Admin.} For the beta, admins serve as maintainers of PrivBench\footnote{\scriptsize This will be revisited if PrivBench scales beyond the current beta version.}. Admins manage all current benchmarking modules, which includes the ability to add, edit, or remove modules. When adding or editing modules, the admin is responsible for providing the source code (Python logic) and requirements files, as well as selecting appropriate dataset mappings. Whenever any changes are made to the collection of modules, the changes are not enforced until the admin publishes the updates. Publishing requires an admin to update the PrivBench version, following standard versioning convention (e.g., v1.1.2). Here, an admin must also provide version notes, which are published on the version history page. Upon publishing updates, users are informed in their submissions panel (as well as via email), and they are prompted to update their submissions to reflect the newest PrivBench version.

Admins are also able to manage datasets, which allows for future updates in which new or updated datasets are used for module execution. Likewise, admins can manage all users of the platform, with the primary capabilities being the deletion of users (for example, spam or stale users), as well as updating daily submission allowances. Finally, the admin panel features direct bridges to the underlying VM, where the admin can monitor, restart, or delete Docker containers and images (i.e., those running the modules). A simple hardware utilization dashboard enables the admin to track VM load.

\paragraph{Versioning.}
As introduced in the admin journey, we have constructed PrivBench to evolve with the inevitable changes that will come in the understanding of how to best evaluate text privatization. As such, PrivBench features a git-style versioning system, which allows for the addition, removal, or updating of component evaluation modules. Once a change (or batch of changes) are published by an admin, PrivBench is upgraded to a new version (following the \texttt{Major.Minor.Patch} convention). Older versions are persisted, including the scores of all submissions under this version. Rollbacks are also possible via the admin panel. A complete account of the \href{https://privbench.com/version-history}{Version History} is publicly viewable.

All platform users are notified via email when a new version of PrivBench is published. In the user's submissions panel, two actions are possible. In the case of a module \textit{deletion}, the user can simply select to update the submission to the newest version (necessitating only a re-calculation of the overall score). However, if a module is updated or added (including dataset changes), the user will be prompted to re-run their submission, or if necessary, provided corresponding newly privatized datasets. This system allows for the continual evolution of PrivBench to adapt with state-of-the-art research, but also for flexibility in the user experience to update submissions if and when desired.


\paragraph{License of use.}
We release the entire PrivBench codebase publicly, which includes the code for both the web application and the underlying benchmarking modules, as well as runner scripts for all tested privatization methods (see next Section \ref{sec:case}). The code repository is released under a GPLv3 license. The repository can be found at: \url{https://github.com/sjmeis/PrivBench}

We are also committed to initially providing PrivBench as a free-to-use resource for researchers and practitioners in the privacy community. Thus, the beta release of PrivBench is provided under a permissive, yet non-commercial CC BY-NC-ND license. Although we impose a daily submission limit (default: 5) to enable fair resource sharing and to manage server load, we are open to requests for higher limits based on need. We make our \href{https://privbench.com/privacy}{Privacy Policy} and \href{https://privbench.com/tos}{Terms of Service} clear.

Finally, we seek to gradually add admins and code contributors to the platform in order to create a diverse and rounded administrative team to foster the growth and sustainability of PrivBench. Members of the community are welcome to contact us if interested at \href{mailto:admin@privbench.com}{admin@privbench.com}.

\section{A Comparative Case Study on Holistic Text Privatization Evaluation}
\label{sec:case}
To showcase the benefits of PrivBench, more specifically its purpose as an enabler of comparability between text-to-text privatization methods, we conduct a case study that evaluates 14 privatization methods across a diverse range of sub-disciplines. 

\paragraph{Methods.}
We select 14 text-to-text privatization methods (29 configurations), which primarily stem from those screened during our literature review.

\setcounter{footnote}{3}
We test methods focusing on entity-based anonymization, including Presidio\footnote{\scriptsize \url{https://github.com/data-privacy-stack/presidio}}, Philter \cite{norgeot2020protected}, Textwash \cite{kleinberg2022textwashautomatedopensource}, and GLiNER \cite{zaratiana-etal-2024-gliner}. These are run with all defaults of the public implementations.

We include several Differential Privacy-based methods, including the word-level 1-Diffractor \cite{10.1145/3643651.3659896}, the token-replacement methods of SanText(+) \cite{yue-etal-2021-differential}, CusText(+) \cite{chen-etal-2023-customized}, DP-Paraphrase \cite{mattern-etal-2022-limits}, DP-Prompt \cite{utpala-etal-2023-locally}, DP-MLM \cite{meisenbacher2024dp}, the document-level ADePT \cite{krishna-etal-2021-adept}, and DP-BART \cite{igamberdiev-habernal-2023-dp}. To ensure a fair evaluation, we utilize these methods with a document-level privacy parameter ($\varepsilon$) equal to the average word length per document per dataset (i.e., $\varepsilon = mean(\text{doc\_length})$), as well as $2\cdot \varepsilon$.

We also include RUPTA \cite{yang-etal-2025-robust}, a method that privatizes texts in a multi-LLM setup, and Privacy Filter\footnote{\scriptsize\url{https://openai.com/index/introducing-openai-privacy-filter/}}, a new model from OpenAI.

With entity-based methods (Presidio, GLiNER, and Privacy Filter), we test two modes: (1) \textit{redact}, where private entities are deleted, and (2) \textit{replace}, where these are placeholdered, e.g., \say{[NAME]}.

\paragraph{Datasets.}
Each PrivBench module is mapped to one or more datasets (Table \ref{tab:modules}). The current datasets used are IMDb \cite{maas-etal-2011-learning}, WikiText \cite{merity2017pointer}, Yelp \cite{10.5555/2969239.2969312,utpala-etal-2023-locally}, PubMedQA \cite{jin-etal-2019-pubmedqa}, TAB \cite{pilan-etal-2022-text}, and Reddit Mental Health \cite{meisenbacher-etal-2025-leveraging}. We take a 1000-row random sample from each dataset. In the case of multiple datasets being mapped to a module, the resulting scores are averaged over all datasets.

\begin{table*}[t!]
\centering
\resizebox{0.89\linewidth}{!}{
\begin{tabular}{l|ccccc|ccccc||cc}
 & ATT & MTI & NNP & PEM & LRO & COH & MAU & SIM & TUT & LVA & $\mu$ 1.0 & $\mu$ 1.1\\ \hline
Presidio (Redact) & 0.0 & 58.9 & 0.1 & 52.0 & 29.0 & 74.3 & 87.0 & \textbf{100} & \textbf{100} & 72.2 & 57.4 & 47.8 \\
Presidio (Replace) & 0.4 & 57.3 & 0.1 & 51.5 & 28.8 & \textbf{100} & 87.7 & \textbf{100} & \textbf{100} & 72.1 & 59.8 & 51.2 \\
Philter & 0.4 & 57.2 & 0.2 & 29.0 & 9.2 & 99.6 & 93.8 & \textbf{100} & 99.9 & 71.8 & 56.1 & 48.3 \\
Textwash & 0.0 & 68.9 & 0.3 & 87.0 & 54.7 & \textbf{100} & 65.2 & \textbf{100} & 99.0 & 70.0 & \textbf{64.5} & 56.1 \\
GLiNER (Redact) & 0.0 & 59.8 & 0.2 & 23.0 & 19.7 & 75.6 & 82.3 & \textbf{100} & \textbf{100} & 73.2 & 53.4 & 46.9 \\
GLiNER (Replace) & 0.2 & 58.2 & 0.2 & 23.0 & 19.8 & \textbf{100} & 91.2 & \textbf{100} & \textbf{100} & 73.2 & 56.6 & 50.2 \\
\hline
1-Diffractor ($\varepsilon$) & 6.5 & 60.1 & 0.1 & 49.5 & 26.0 & 34.0 & 62.3 & 94.6 & 98.8 & 70.0 & 50.2 & 42.2 \\
1-Diffractor ($2\cdot\varepsilon$) & 0.4 & 60.1 & 0.1 & 59.5 & 20.5 & 43.4 & 18.5 & 98.6 & 99.5 & 71.3 & 47.2 & 42.2 \\
SanText ($\varepsilon$) & 32.1 & 68.5 & 31.5 & 97.0 & 29.0 & 1.0 & 5.3 & 65.2 & 77.2 & 69.3 & 47.6 & 44.1 \\
SanText ($2\cdot\varepsilon$) & \textbf{37.9} & 67.0 & 20.3 & 89.0 & 20.2 & 1.7 & 14.6 & 72.4 & 83.4 & 69.3 & 47.6 & 42.8 \\
SanText+ ($\varepsilon$) & 16.3 & 60.8 & 0.1 & 47.5 & 35.3 & 7.7 & 21.4 & 88.8 & 95.3 & 69.3 & 44.2 & 49.3 \\
SanText+ ($2\cdot\varepsilon$) & 13.0 & 59.7 & 0.2 & 46.0 & 25.5 & 9.8 & 24.4 & 90.2 & 96.3 & 69.2 & 43.4 & 39.1 \\
CusText ($\varepsilon$) & 23.5 & 62.0 & 0.8 & 93.0 & 77.8 & 1.8 & 6.7 & 78.4 & 84.7 & 69.3 & 49.8 & 45.7 \\
CusText ($2\cdot\varepsilon$) & 23.8 & 64.8 & 0.6 & 92.0 & 78.0 & 1.9 & 5.9 & 79.4 & 85.1 & 69.3 & 50.1 & 46.2 \\
CusText+ ($\varepsilon$) & 28.7 & 66.1 & 0.3 & 93.5 & 86.2 & 4.8 & 10.6 & 81.3 & 92.4 & 69.3 & 53.3 & 49.7 \\
CusText+ ($2\cdot\varepsilon$) & 30.1 & 66.7 & 0.2 & 93.0 & 83.7 & 5.4 & 10.4 & 82.4 & 93.1 & 69.3 & 53.4 & 49.8 \\
DP-Paraphrase ($\varepsilon$) & 32.3 & 70.3 & 9.1 & 97.0 & 88.8 & 21.8 & 7.4 & 65.7 & 80.4 & 76.4 & 54.9 & 54.2 \\
DP-Paraphrase ($2\cdot\varepsilon$) & 35.9 & 67.8 & 9.5 & 96.0 & 88.8 & 36.0 & 10.9 & 64.1 & 78.5 & \textbf{78.1} & 56.6 & 57.8 \\
DP-Prompt ($\varepsilon$) & 31.6 & 69.2 & 21.3 & 92.0 & 65.3 & 2.0 & 15.2 & 78.1 & 86.8 & 59.0 & 52.0 & 47.9 \\
DP-Prompt ($2\cdot\varepsilon$) & 29.6 & 69.9 & 4.5 & 83.0 & 33.5 & 16.7 & 25.9 & 83.2 & 91.5 & 50.4 & 48.8 & 42.3 \\
DP-MLM ($\varepsilon$) & 23.5 & 64.7 & 34.8 & \textbf{100} & \textbf{100} & 1.8 & 4.8 & 53.9 & 78.0 & 71.2 & 53.3 & 53.4 \\
DP-MLM ($2\cdot\varepsilon$) & 21.7 & 64.2 & 31.5 & \textbf{100} & \textbf{100} & 1.9 & 4.3 & 54.3 & 76.6 & 71.5 & 52.6 & 52.7 \\
ADePT ($\varepsilon$) & 31.1 & 61.1 & 42.1 & 98.0 & \textbf{100} & \textbf{100} & 2.7 & 58.1 & 74.9 & 28.8 & 59.7 & \textbf{62.6} \\
ADePT ($2\cdot\varepsilon$) & 30.1 & 74.3 & 39.6 & 97.5 & 56.8 & \textbf{100} & 2.3 & 61.4 & 74.9 & 33.2 & 57.0 & 58.4 \\
DP-BART ($\varepsilon$) & 28.2 & \textbf{75.6} & \textbf{47.9} & \textbf{100} & \textbf{100} & 0.1 & 5.0 & 52.3 & 77.5 & 64.2 & 55.1 & 56.2 \\
DP-BART ($2\cdot\varepsilon$) & 28.7 & 68.6 & 46.9 & \textbf{100} & \textbf{100} & 0.1 & 5.3 & 52.2 & 76.2 & 64.1 & 54.2 & 55.0 \\
\hline
RUPTA & 2.4 & 57.6 & 0.2 & 49.0 & 16.8 & \textbf{100} & 69.4 & \textbf{100} & 99.9 & 73.9 & 56.9 & 50.1 \\
Privacy Filter (Redact) & 0.0 & 58.2 & 0.1 & 10.0 & 18.8 & 96.1 & \textbf{99.9} & \textbf{100} & \textbf{100} & 72.4 & 55.6 & 49.4 \\
Privacy Filter (Replace) & 0.0 & 57.9 & 0.1 & 10.0 & 19.0 & \textbf{100} & 99.8 & \textbf{100} & \textbf{100} & 72.2 & 55.9 & 49.9 \\
\end{tabular}
}
\caption{Case study results. All scores are out of 100 ($\uparrow$), and the final score ($\mu$) is the mean of all 10 module scores (v1.0.0), or the remaining seven modules in v1.1.0 (i.e., excluding PEM, MAU, and SIM, see Section \ref{sec:remove}). \textbf{Bold} scores denote the best per column. Module abbreviations are from Table \ref{tab:modules}.}
\label{tab:results}
\end{table*}

\paragraph{Results.}
The case study results are presented in Table \ref{tab:results}, showing how our holistic approach enables a comprehensive as well as nuanced view of privatization performance. While DP-based methods are strong in ensuring privacy protections, they are not in utility metrics. Conversely, entity-based methods excel in utility but are poor in achieving meaningful privacy protection. The utility-preserving capabilities, however, dominate the privacy-utility balance, leading to higher overall scores.

These results highlight the great difficulty text privatization approaches have in preserving utility alongside strong privacy. Indeed, the two sides of Table \ref{tab:results} confirm that few methods can achieve consistently similar results in both. Thus, a crucial research gap is highlighted, namely the design of methods that empirically protect privacy, but not at the cost of extreme utility degradation. We propose PrivBench as a facilitator in tackling this challenge. 

\paragraph{From case study to PrivBench v1.1.0.}
\label{sec:remove}
To internally verify the robustness and fairness of the PrivBench evaluation modules, we checked for inter-module redundancies, using the results of our case study as data points. With the scores of Table \ref{tab:results}, we calculated the Spearman's rank correlation coefficient ($r_s$) between all modules. In particular, we calculated correlations separately between the five utility and five privacy modules. This analysis revealed five highly correlated modules ($r_s > 0.8, p < 0.05$), namely: SIM $\leftrightarrow$ TUT (0.96),  MAU $\leftrightarrow$ TUT (0.94), MAU $\leftrightarrow$ SIM (0.93), PEM $\leftrightarrow$ LRO (0.90), and NNP $\leftrightarrow$ PEM (0.85).

To address these redundancies and mitigate the risk of multiply rewarding the same evaluation signal, we propose an updated PrivBench v1.1.0 that removes SIM, MAU, and PEM, thus leaving three utility modules and four privacy modules. Note: in v1.1.0, PEM is still used as the basis for LRO.

The updated scores for v1.1.0 are also provided in Table \ref{tab:results}. As shown there, the module validation (i.e., redundancy removal) shifts the landscape in terms of prevailing scores. We find that DP-based methods are able to reach more favorable balances, whereas the difficulties of non-DP methods in providing empirical privacy protections are now more prominently emphasized. Nevertheless, the overall score distributions still lend concrete evidence to the fact that text-to-text privatization has a significant challenge ahead in optimizing the delicate balance between privacy and utility.

\section{Conclusion}
We introduce PrivBench, a fully deployed, public, and free-to-use benchmarking platform motivated by the holistic and modular evaluation of text-to-text privatization methods. The base platform is released with 10 benchmarking modules in v1.0.0 and a reduced set of seven in v1.1.0. PrivBench creates a space for researchers and practitioners working in text privacy to collaborate in a competitive yet fair environment, and concurrently, for motivated administrators to ensure PrivBench’s upkeep and to guide its future. Our case study showcases the opportunity of PrivBench to promote better comparability and higher transparency of text privatization effectiveness. In this spirit, we hope that PrivBench helps to push the boundaries of text privatization, and that it continues to evolve as the understanding of the evaluation thereof advances. 

\section*{Limitations}
As with the general evaluation of \textit{text privatization}, PrivBench is limited to the current understanding of how to best evaluate \say{good} technical approaches to privacy protection in textual data. We mitigate this limitation via our modular approach, which holistically captures different aspects of text privatization evaluation. As such, PrivBench can be modified and updated as more evaluation strategies are developed (or suggested) by the community. We remain committed to facilitating continuous improvements to the modular benchmark, to build on the basis we establish with this first iteration.

We decided to strike a balance between \textit{comprehensive} and \textit{efficient} evaluation. As such, evaluation datasets (those associated with modules) are capped at 1000 rows/texts per module. While this eases the burden and turnaround time for users of the platform, we acknowledge that larger datasets would lead to more robust scoring. This remains a point for future work, especially if PrivBench is migrated to a host with higher resource capacities.

In the weighting of the different evaluation modules, we opted for a simple equal weighting of all the modules for the calculation of the overall score per submission. Future work may improve this simpler scheme by assigning relative weights, which could be obtained either via automatic means or through the modeling of human preferences.

We also acknowledge that due to the nature of the platform design, we cannot evaluate the computational efficiency of proposed text privatization methods, which would also be a key evaluation criterion, particularly for practical applicability.

Although we conducted our case study to validate the function and capabilities of PrivBench, we did not run user studies to validate its user-friendliness and general UX. This would represent a logical next step in the development of PrivBench, particularly as users begin to adopt the platform.

\section*{Ethical Considerations}
We confirm that all underlying datasets that comprise the current PrivBench are used in accordance with their original (open-source) licenses, namely:

\begin{itemize}
    \itemsep -0.5em
    \item IMDb: CC BY 4.0
    \item WikiText: CC BY-SA 4.0
    \item Yelp: \href{https://www.yelp.com/brand/dataset_terms}{non-commercial}
    \item PubMedQA: MIT
    \item TAB: MIT
    \item Reddit: \href{https://redditinc.com/policies/developer-terms}{non-commercial}
\end{itemize}

In solely using public research datasets, we mitigate and discourage adversarial or malicious use of PrivBench. As a result of this structure, we do not elicit or ask for the submission of proprietary or private data to participate in the benchmark. Proper use of the website involves only submitting privatized versions of the provided, publicly available datasets. We explicitly advise users against uploading personally identifiable information, proprietary content, or sensitive private data in lieu of the provided datasets. Any upload of such data is strictly at the user’s sole discretion and risk.

We take further measures to prevent malicious use of our website, including strong access controls, bot prevention, and following standard secure web development practices. In addition, we invite and appreciate any reports of issues or potential security vulnerabilities. The Code of Conduct and contribution guidelines are found in our repository. 

Finally, we make our \href{https://privbench.com/privacy}{Privacy Policy} and \href{https://privbench.com/tos}{Terms of Service} transparent and clear, and we will work to keep these updated as PrivBench evolves.

\section*{Acknowledgments}
We extend a heartfelt thank you to the following people for making PrivBench possible. Firstly, to Han My Do, Simeon Ivanov, Souraya Kharfı, and Tobias Unterhauser for the foundation. To James Flemings, Timour Igamberdiev, and Ibrahim Baroud for the support in the initial literature review and brainstorming. And to Muhammad Jibran for the systems support and the rest of sebis for the general support. We appreciate you all very much!

\bibliography{custom,slr}

@inproceedings{habernal-etal-2023-privacy,
    title = "Privacy-Preserving Natural Language Processing",
    author = "Habernal, Ivan  and
      Mireshghallah, Fatemehsadat  and
      Thaine, Patricia  and
      Ghanavati, Sepideh  and
      Feyisetan, Oluwaseyi",
    editor = "Zanzotto, Fabio Massimo  and
      Pradhan, Sameer",
    booktitle = "Proceedings of the 17th Conference of the European Chapter of the Association for Computational Linguistics: Tutorial Abstracts",
    month = may,
    year = "2023",
    address = "Dubrovnik, Croatia",
    publisher = "Association for Computational Linguistics",
    url = "https://aclanthology.org/2023.eacl-tutorials.6/",
    doi = "10.18653/v1/2023.eacl-tutorials.6",
    pages = "27--30"
}

@INPROCEEDINGS{11247969,
  author={Deußer, Tobias and Sparrenberg, Lorenz and Berger, Armin and Hahnbück, Max and Bauckhage, Christian and Sifa, Rafet},
  booktitle={2025 IEEE 12th International Conference on Data Science and Advanced Analytics (DSAA)}, 
  title={A Survey on Current Trends and Recent Advances in Text Anonymization}, 
  year={2025},
  volume={},
  number={},
  pages={1-9},
  doi={10.1109/DSAA65442.2025.11247969}}

@inproceedings{mattern-etal-2022-limits,
    title = "The Limits of Word Level Differential Privacy",
    author = "Mattern, Justus  and
      Weggenmann, Benjamin  and
      Kerschbaum, Florian",
    editor = "Carpuat, Marine  and
      de Marneffe, Marie-Catherine  and
      Meza Ruiz, Ivan Vladimir",
    booktitle = "Findings of the Association for Computational Linguistics: NAACL 2022",
    month = jul,
    year = "2022",
    address = "Seattle, United States",
    publisher = "Association for Computational Linguistics",
    url = "https://aclanthology.org/2022.findings-naacl.65",
    doi = "10.18653/v1/2022.findings-naacl.65",
    pages = "867--881"
}

@inproceedings{ren-etal-2025-measure,
    title = "How do we measure privacy in text? A survey of text anonymization metrics",
    author = "Ren, Yaxuan  and
      Ramesh, Krithika  and
      Yao, Yaxing  and
      Field, Anjalie",
    editor = "Inui, Kentaro  and
      Sakti, Sakriani  and
      Wang, Haofen  and
      Wong, Derek F.  and
      Bhattacharyya, Pushpak  and
      Banerjee, Biplab  and
      Ekbal, Asif  and
      Chakraborty, Tanmoy  and
      Singh, Dhirendra Pratap",
    booktitle = "Proceedings of the 14th International Joint Conference on Natural Language Processing and the 4th Conference of the Asia-Pacific Chapter of the Association for Computational Linguistics",
    month = dec,
    year = "2025",
    address = "Mumbai, India",
    publisher = "The Asian Federation of Natural Language Processing and The Association for Computational Linguistics",
    url = "https://aclanthology.org/2025.findings-ijcnlp.94/",
    pages = "1532--1544",
    ISBN = "979-8-89176-303-6"
}

@inproceedings{loiseau-etal-2025-tau,
    title = "Tau-Eval: A Unified Evaluation Framework for Useful and Private Text Anonymization",
    author = "Loiseau, Gabriel  and
      Sileo, Damien  and
      Riquet, Damien  and
      Meyer, Maxime  and
      Tommasi, Marc",
    editor = {Habernal, Ivan  and
      Schulam, Peter  and
      Tiedemann, J{\"o}rg},
    booktitle = "Proceedings of the 2025 Conference on Empirical Methods in Natural Language Processing: System Demonstrations",
    month = nov,
    year = "2025",
    address = "Suzhou, China",
    publisher = "Association for Computational Linguistics",
    url = "https://aclanthology.org/2025.emnlp-demos.16/",
    doi = "10.18653/v1/2025.emnlp-demos.16",
    pages = "216--227",
    ISBN = "979-8-89176-334-0"
}

@article{pilan-etal-2022-text,
    title = "The Text Anonymization Benchmark ({TAB}): A Dedicated Corpus and Evaluation Framework for Text Anonymization",
    author = "Pil{\'a}n, Ildik{\'o}  and
      Lison, Pierre  and
      {\O}vrelid, Lilja  and
      Papadopoulou, Anthi  and
      S{\'a}nchez, David  and
      Batet, Montserrat",
    journal = "Computational Linguistics",
    volume = "48",
    number = "4",
    month = dec,
    year = "2022",
    address = "Cambridge, MA",
    publisher = "MIT Press",
    url = "https://aclanthology.org/2022.cl-4.19/",
    doi = "10.1162/coli_a_00458",
    pages = "1053--1101"
}

@inproceedings{klymenko-etal-2022-differential,
    title = "Differential Privacy in Natural Language Processing: The Story So Far",
    author = "Klymenko, Oleksandra  and
      Meisenbacher, Stephen  and
      Matthes, Florian",
    editor = "Feyisetan, Oluwaseyi  and
      Ghanavati, Sepideh  and
      Thaine, Patricia  and
      Habernal, Ivan  and
      Mireshghallah, Fatemehsadat",
    booktitle = "Proceedings of the Fourth Workshop on Privacy in Natural Language Processing",
    month = jul,
    year = "2022",
    address = "Seattle, United States",
    publisher = "Association for Computational Linguistics",
    url = "https://aclanthology.org/2022.privatenlp-1.1/",
    doi = "10.18653/v1/2022.privatenlp-1.1",
    pages = "1--11"
}

@inproceedings{yue-etal-2023-synthetic,
    title = "Synthetic Text Generation with Differential Privacy: A Simple and Practical Recipe",
    author = "Yue, Xiang  and
      Inan, Huseyin  and
      Li, Xuechen  and
      Kumar, Girish  and
      McAnallen, Julia  and
      Shajari, Hoda  and
      Sun, Huan  and
      Levitan, David  and
      Sim, Robert",
    editor = "Rogers, Anna  and
      Boyd-Graber, Jordan  and
      Okazaki, Naoaki",
    booktitle = "Proceedings of the 61st Annual Meeting of the Association for Computational Linguistics (Volume 1: Long Papers)",
    month = jul,
    year = "2023",
    address = "Toronto, Canada",
    publisher = "Association for Computational Linguistics",
    url = "https://aclanthology.org/2023.acl-long.74/",
    doi = "10.18653/v1/2023.acl-long.74",
    pages = "1321--1342"
}

@inproceedings{wang-etal-2018-glue,
    title = "{GLUE}: A Multi-Task Benchmark and Analysis Platform for Natural Language Understanding",
    author = "Wang, Alex  and
      Singh, Amanpreet  and
      Michael, Julian  and
      Hill, Felix  and
      Levy, Omer  and
      Bowman, Samuel",
    editor = "Linzen, Tal  and
      Chrupa{\l}a, Grzegorz  and
      Alishahi, Afra",
    booktitle = "Proceedings of the 2018 {EMNLP} Workshop {B}lackbox{NLP}: Analyzing and Interpreting Neural Networks for {NLP}",
    month = nov,
    year = "2018",
    address = "Brussels, Belgium",
    publisher = "Association for Computational Linguistics",
    url = "https://aclanthology.org/W18-5446/",
    doi = "10.18653/v1/W18-5446",
    pages = "353--355"
}

@inproceedings{hendrycks2021measuringmassivemultitasklanguage,
  title={Measuring Massive Multitask Language Understanding},
  author={Hendrycks, Dan and Burns, Collin and Basart, Steven and Zou, Andy and Mazeika, Mantas and Song, Dawn and Steinhardt, Jacob},
  booktitle={International Conference on Learning Representations},
  year={2021},
  url={https://openreview.net/forum?id=d7KBjmI3GmQ}
}

@article{srivastava2023beyond,
  title={Beyond the imitation game: Quantifying and extrapolating the capabilities of language models},
  author={Srivastava, Aarohi and Rastogi, Abhinav and Rao, Abhishek and Shoeb, Abu Awal Md and Abid, Abubakar and Fisch, Adam and Brown, Adam R and Santoro, Adam and Gupta, Aditya and Garriga-Alonso, Adri{\`a} and others},
  journal={Transactions on machine learning research},
  year={2023},
  url={https://openreview.net/forum?id=uyTL5Bvosj}
}

@article{liangholistic,
  title={Holistic Evaluation of Language Models},
  author={Liang, Percy and Bommasani, Rishi and Lee, Tony and Tsipras, Dimitris and Soylu, Dilara and Yasunaga, Michihiro and Zhang, Yian and Narayanan, Deepak and Wu, Yuhuai and Kumar, Ananya and others},
  journal={Transactions on Machine Learning Research},
  url={https://openreview.net/forum?id=iO4LZibEqW},
  year={2022}
}

@article{sousa2023keep,
  title={How to keep text private? A systematic review of deep learning methods for privacy-preserving natural language processing},
  author={Sousa, Samuel and Kern, Roman},
  journal={Artificial Intelligence Review},
  volume={56},
  number={2},
  pages={1427--1492},
  year={2023},
  publisher={Springer},
  doi={10.1007/s10462-022-10204-6}
}

@inproceedings{li-etal-2024-privlm,
    title = "{P}riv{LM}-Bench: A Multi-level Privacy Evaluation Benchmark for Language Models",
    author = "Li, Haoran  and
      Guo, Dadi  and
      Li, Donghao  and
      Fan, Wei  and
      Hu, Qi  and
      Liu, Xin  and
      Chan, Chunkit  and
      Yao, Duanyi  and
      Yao, Yuan  and
      Song, Yangqiu",
    editor = "Ku, Lun-Wei  and
      Martins, Andre  and
      Srikumar, Vivek",
    booktitle = "Proceedings of the 62nd Annual Meeting of the Association for Computational Linguistics (Volume 1: Long Papers)",
    month = aug,
    year = "2024",
    address = "Bangkok, Thailand",
    publisher = "Association for Computational Linguistics",
    url = "https://aclanthology.org/2024.acl-long.4/",
    doi = "10.18653/v1/2024.acl-long.4",
    pages = "54--73"
}

@article{shao2024privacylens,
  title={{PrivacyLens}: Evaluating privacy norm awareness of language models in action},
  author={Shao, Yijia and Li, Tianshi and Shi, Weiyan and Liu, Yanchen and Yang, Diyi},
  journal={Advances in Neural Information Processing Systems},
  volume={37},
  pages={89373--89407},
  year={2024},
  doi={10.52202/079017-2837}
}

@article{zhu2024privauditor,
  title={{PrivAuditor}: Benchmarking data protection vulnerabilities in {LLM} adaptation techniques},
  author={Zhu, Derui and Chen, Dingfan and Wu, Xiongfei and Geng, Jiahui and Li, Zhuo and Grossklags, Jens and Ma, Lei},
  journal={Advances in Neural Information Processing Systems},
  volume={37},
  pages={9668--9689},
  year={2024},
  doi={10.52202/079017-0308}
}

@inproceedings{meisenbacher-etal-2024-comparative,
    title = "A Comparative Analysis of Word-Level Metric Differential Privacy: Benchmarking the Privacy-Utility Trade-off",
    author = "Meisenbacher, Stephen  and
      Nandakumar, Nihildev  and
      Klymenko, Alexandra  and
      Matthes, Florian",
    editor = "Calzolari, Nicoletta  and
      Kan, Min-Yen  and
      Hoste, Veronique  and
      Lenci, Alessandro  and
      Sakti, Sakriani  and
      Xue, Nianwen",
    booktitle = "Proceedings of the 2024 Joint International Conference on Computational Linguistics, Language Resources and Evaluation (LREC-COLING 2024)",
    month = may,
    year = "2024",
    address = "Torino, Italia",
    publisher = "ELRA and ICCL",
    url = "https://aclanthology.org/2024.lrec-main.16/",
    pages = "174--185"
}

@inproceedings{igamberdiev-etal-2022-dp,
    title = "{DP}-Rewrite: Towards Reproducibility and Transparency in Differentially Private Text Rewriting",
    author = "Igamberdiev, Timour  and
      Arnold, Thomas  and
      Habernal, Ivan",
    editor = "Calzolari, Nicoletta  and
      Huang, Chu-Ren  and
      Kim, Hansaem  and
      Pustejovsky, James  and
      Wanner, Leo  and
      Choi, Key-Sun  and
      Ryu, Pum-Mo  and
      Chen, Hsin-Hsi  and
      Donatelli, Lucia  and
      Ji, Heng  and
      Kurohashi, Sadao  and
      Paggio, Patrizia  and
      Xue, Nianwen  and
      Kim, Seokhwan  and
      Hahm, Younggyun  and
      He, Zhong  and
      Lee, Tony Kyungil  and
      Santus, Enrico  and
      Bond, Francis  and
      Na, Seung-Hoon",
    booktitle = "Proceedings of the 29th International Conference on Computational Linguistics",
    month = oct,
    year = "2022",
    address = "Gyeongju, Republic of Korea",
    publisher = "International Committee on Computational Linguistics",
    url = "https://aclanthology.org/2022.coling-1.258/",
    pages = "2927--2933"
}

@misc{sun2025synbenchbenchmarkdifferentiallyprivate,
      title={{SynBench}: A Benchmark for Differentially Private Text Generation}, 
      author={Yidan Sun and Viktor Schlegel and Srinivasan Nandakumar and Iqra Zahid and Yuping Wu and Yulong Wu and Hao Li and Jie Zhang and Warren Del-Pinto and Goran Nenadic and Siew Kei Lam and Anil Anthony Bharath},
      year={2025},
      eprint={2509.14594},
      archivePrefix={arXiv},
      primaryClass={cs.AI},
      url={https://arxiv.org/abs/2509.14594}, 
      doi={10.48550/arXiv.2509.14594}
}

@book{10.5555/2994449,
author = {Kitchenham, Barbara Ann and Budgen, David and Brereton, Pearl},
title = {Evidence-Based Software Engineering and Systematic Reviews},
year = {2015},
isbn = {1482228653},
publisher = {Chapman \& Hall/CRC},
doi={10.5555/2994449}
}

@article{tong2025inferdpt,
  author={Tong, Meng and Chen, Kejiang and Zhang, Jie and Qi, Yuang and Zhang, Weiming and Yu, Nenghai and Zhang, Tianwei and Zhang, Zhikun},
  journal={IEEE Transactions on Dependable and Secure Computing}, 
  title={{InferDPT}: Privacy-Preserving Inference for Closed-Box Large Language Models}, 
  year={2025},
  volume={22},
  number={5},
  pages={4625-4640},
  doi={10.1109/TDSC.2025.3550389}}

@inproceedings{vinod2025invisibleink,
  title={{InvisibleInk}: High-Utility and Low-Cost Text Generation with Differential Privacy},
  author={Vinod, Vishnu and Pillutla, Krishna and Thakurta, Abhradeep Guha},
  booktitle={The Thirty-ninth Annual Conference on Neural Information Processing Systems},
  year={2025},
  url={https://openreview.net/forum?id=B4NT8TexNS}
}

@inproceedings{wang2024knowledgesg,
    title = "{K}nowledge{SG}: Privacy-Preserving Synthetic Text Generation with Knowledge Distillation from Server",
    author = "Wang, WenHao  and
      Liang, Xiaoyu  and
      Ye, Rui  and
      Chai, Jingyi  and
      Chen, Siheng  and
      Wang, Yanfeng",
    editor = "Al-Onaizan, Yaser  and
      Bansal, Mohit  and
      Chen, Yun-Nung",
    booktitle = "Proceedings of the 2024 Conference on Empirical Methods in Natural Language Processing",
    month = nov,
    year = "2024",
    address = "Miami, Florida, USA",
    publisher = "Association for Computational Linguistics",
    url = "https://aclanthology.org/2024.emnlp-main.438/",
    doi = "10.18653/v1/2024.emnlp-main.438",
    pages = "7677--7695"
}

@inproceedings{carranza2024synthetic,
    title = "Synthetic Query Generation for Privacy-Preserving Deep Retrieval Systems using Differentially Private Language Models",
    author = "Carranza, Aldo  and
      Farahani, Rezsa  and
      Ponomareva, Natalia  and
      Kurakin, Alexey  and
      Jagielski, Matthew  and
      Nasr, Milad",
    editor = "Duh, Kevin  and
      Gomez, Helena  and
      Bethard, Steven",
    booktitle = "Proceedings of the 2024 Conference of the North American Chapter of the Association for Computational Linguistics: Human Language Technologies (Volume 1: Long Papers)",
    month = jun,
    year = "2024",
    address = "Mexico City, Mexico",
    publisher = "Association for Computational Linguistics",
    url = "https://aclanthology.org/2024.naacl-long.217/",
    doi = "10.18653/v1/2024.naacl-long.217",
    pages = "3920--3930"
}

@article{uchendu2023attribution,
author = {Uchendu, Adaku and Le, Thai and Lee, Dongwon},
title = {Attribution and Obfuscation of Neural Text Authorship: A Data Mining Perspective},
year = {2023},
issue_date = {June 2023},
publisher = {Association for Computing Machinery},
address = {New York, NY, USA},
volume = {25},
number = {1},
issn = {1931-0145},
url = {https://doi.org/10.1145/3606274.3606276},
doi = {10.1145/3606274.3606276},
journal = {SIGKDD Explor. Newsl.},
month = jul,
pages = {1–18},
numpages = {18}
}

@inproceedings{xie2024differentially,
author = {Xie, Chulin and Lin, Zinan and Backurs, Arturs and Gopi, Sivakanth and Yu, Da and Inan, Huseyin and Nori, Harsha and Jiang, Haotian and Zhang, Huishuai and Lee, Yin Tat and Li, Bo and Yekhanin, Sergey},
title = {Differentially private synthetic data via foundation model {APIs} 2: text},
year = {2024},
publisher = {JMLR.org},
booktitle = {Proceedings of the 41st International Conference on Machine Learning},
articleno = {2243},
numpages = {30},
location = {Vienna, Austria},
series = {ICML'24},
url={https://dl.acm.org/doi/10.5555/3692070.3694313}
}

@inproceedings{mattern2022differentially,
    title = "Differentially Private Language Models for Secure Data Sharing",
    author = {Mattern, Justus  and
      Jin, Zhijing  and
      Weggenmann, Benjamin  and
      Sch{\"o}lkopf, Bernhard  and
      Sachan, Mrinmaya},
    editor = "Goldberg, Yoav  and
      Kozareva, Zornitsa  and
      Zhang, Yue",
    booktitle = "Proceedings of the 2022 Conference on Empirical Methods in Natural Language Processing",
    month = dec,
    year = "2022",
    address = "Abu Dhabi, United Arab Emirates",
    publisher = "Association for Computational Linguistics",
    url = "https://aclanthology.org/2022.emnlp-main.323/",
    doi = "10.18653/v1/2022.emnlp-main.323",
    pages = "4860--4873"
}

@inproceedings{vats2024recovering,
  author={Vats, Arpita and Liu, Zhe and Su, Peng and Paul, Debjyoti and Ma, Yingyi and Pang, Yutong and Ahmed, Zeeshan and Kalinli, Ozlem},
  booktitle={ICASSP 2024 - 2024 IEEE International Conference on Acoustics, Speech and Signal Processing (ICASSP)}, 
  title={Recovering from Privacy-Preserving Masking with Large Language Models}, 
  year={2024},
  volume={},
  number={},
  pages={10771-10775},
  doi={10.1109/ICASSP48485.2024.10448234}}

@inproceedings{qu2021natural,
author = {Qu, Chen and Kong, Weize and Yang, Liu and Zhang, Mingyang and Bendersky, Michael and Najork, Marc},
title = {Natural Language Understanding with Privacy-Preserving BERT},
year = {2021},
isbn = {9781450384469},
publisher = {Association for Computing Machinery},
address = {New York, NY, USA},
url = {https://doi.org/10.1145/3459637.3482281},
doi = {10.1145/3459637.3482281},
booktitle = {Proceedings of the 30th ACM International Conference on Information \& Knowledge Management},
pages = {1488–1497},
numpages = {10},
location = {Virtual Event, Queensland, Australia},
series = {CIKM '21}
}

@inproceedings{awon2025clusant,
    title = "{C}lu{S}an{T}: Differentially Private and Semantically Coherent Text Sanitization",
    author = "Awon, Ahmed Musa  and
      Lu, Yun  and
      Potka, Shera  and
      Thomo, Alex",
    editor = "Chiruzzo, Luis  and
      Ritter, Alan  and
      Wang, Lu",
    booktitle = "Proceedings of the 2025 Conference of the Nations of the Americas Chapter of the Association for Computational Linguistics: Human Language Technologies (Volume 1: Long Papers)",
    month = apr,
    year = "2025",
    address = "Albuquerque, New Mexico",
    publisher = "Association for Computational Linguistics",
    url = "https://aclanthology.org/2025.naacl-long.187/",
    doi = "10.18653/v1/2025.naacl-long.187",
    pages = "3676--3693",
    ISBN = "979-8-89176-189-6"
}

@inproceedings{flemings2024differentially_private_knowledge,
    title = "Differentially Private Knowledge Distillation via Synthetic Text Generation",
    author = "Flemings, James  and
      Annavaram, Murali",
    editor = "Ku, Lun-Wei  and
      Martins, Andre  and
      Srikumar, Vivek",
    booktitle = "Findings of the Association for Computational Linguistics: ACL 2024",
    month = aug,
    year = "2024",
    address = "Bangkok, Thailand",
    publisher = "Association for Computational Linguistics",
    url = "https://aclanthology.org/2024.findings-acl.769/",
    doi = "10.18653/v1/2024.findings-acl.769",
    pages = "12957--12968",
}

@inproceedings{flemings2024differentially_private_next_token,
    title = "Differentially Private Next-Token Prediction of Large Language Models",
    author = "Flemings, James  and
      Razaviyayn, Meisam  and
      Annavaram, Murali",
    editor = "Duh, Kevin  and
      Gomez, Helena  and
      Bethard, Steven",
    booktitle = "Proceedings of the 2024 Conference of the North American Chapter of the Association for Computational Linguistics: Human Language Technologies (Volume 1: Long Papers)",
    month = jun,
    year = "2024",
    address = "Mexico City, Mexico",
    publisher = "Association for Computational Linguistics",
    url = "https://aclanthology.org/2024.naacl-long.247/",
    doi = "10.18653/v1/2024.naacl-long.247",
    pages = "4390--4404",
}

@inproceedings{higashi2025differential,
author = {Higashi, Takuya and Nakai, Tsunato},
title = {Is Differential Privacy-Enhanced Parameter-Efficient Fine-Tuning Effective for Large Language Models?},
year = {2025},
isbn = {9798400714153},
publisher = {Association for Computing Machinery},
address = {New York, NY, USA},
url = {https://doi.org/10.1145/3709018.3736329},
doi = {10.1145/3709018.3736329},
booktitle = {Proceedings of the Workshop on Privacy in Large Language Models (LLM) and Natural Language Processing (NLP) 2025},
pages = {1–12},
numpages = {12},
location = {
},
series = {LM-SHIELD '25}
}

@inproceedings{meisenbacher2024spend,
author = {Meisenbacher, Stephen and Lee, Chaeeun Joy and Matthes, Florian},
title = {Spend Your Budget Wisely: Towards an Intelligent Distribution of the Privacy Budget in Differentially Private Text Rewriting},
year = {2025},
isbn = {9798400714764},
publisher = {Association for Computing Machinery},
address = {New York, NY, USA},
url = {https://doi.org/10.1145/3714393.3726504},
doi = {10.1145/3714393.3726504},
booktitle = {Proceedings of the Fifteenth ACM Conference on Data and Application Security and Privacy},
pages = {84–95},
numpages = {12},
location = {Pittsburgh, PA, USA},
series = {CODASPY '25}
}

@inproceedings{meisenbacher2024thinking,
    title = "Thinking Outside of the Differential Privacy Box: A Case Study in Text Privatization with Language Model Prompting",
    author = "Meisenbacher, Stephen  and
      Matthes, Florian",
    editor = "Al-Onaizan, Yaser  and
      Bansal, Mohit  and
      Chen, Yun-Nung",
    booktitle = "Proceedings of the 2024 Conference on Empirical Methods in Natural Language Processing",
    month = nov,
    year = "2024",
    address = "Miami, Florida, USA",
    publisher = "Association for Computational Linguistics",
    url = "https://aclanthology.org/2024.emnlp-main.324/",
    doi = "10.18653/v1/2024.emnlp-main.324",
    pages = "5656--5665"
}

@inproceedings{turan2022adapting,
    title = "Adapting Language Models When Training on Privacy-Transformed Data",
    author = "Turan, Tugtekin  and
      Klakow, Dietrich  and
      Vincent, Emmanuel  and
      Jouvet, Denis",
    editor = "Calzolari, Nicoletta  and
      B{\'e}chet, Fr{\'e}d{\'e}ric  and
      Blache, Philippe  and
      Choukri, Khalid  and
      Cieri, Christopher  and
      Declerck, Thierry  and
      Goggi, Sara  and
      Isahara, Hitoshi  and
      Maegaard, Bente  and
      Mariani, Joseph  and
      Mazo, H{\'e}l{\`e}ne  and
      Odijk, Jan  and
      Piperidis, Stelios",
    booktitle = "Proceedings of the Thirteenth Language Resources and Evaluation Conference",
    month = jun,
    year = "2022",
    address = "Marseille, France",
    publisher = "European Language Resources Association",
    url = "https://aclanthology.org/2022.lrec-1.465/",
    pages = "4367--4373"
}

@inproceedings{li2023beyond,
  title={Beyond Gradient and Priors in Privacy Attacks: Leveraging Pooler Layer Inputs of Language Models in Federated Learning},
  author={Li, Jianwei and Liu, Sheng and Lei, Qi},
  booktitle={International Workshop on Federated Learning in the Age of Foundation Models in Conjunction with NeurIPS 2023},
  year={2023},
  url={https://openreview.net/forum?id=H0inHCV05c}
}

@inproceedings{yu2021differentially,
  title={Differentially Private Fine-tuning of Language Models},
  author={Yu, Da and Naik, Saurabh and Backurs, Arturs and Gopi, Sivakanth and Inan, Huseyin A and Kamath, Gautam and Kulkarni, Janardhan and Lee, Yin Tat and Manoel, Andre and Wutschitz, Lukas and others},
  booktitle={International Conference on Learning Representations},
  year={2022},
  url={https://openreview.net/forum?id=Q42f0dfjECO}
}

@inproceedings{huang2024private,
    title = "Private Language Models via Truncated Laplacian Mechanism",
    author = "Huang, Tianhao  and
      Yang, Tao  and
      Habernal, Ivan  and
      Hu, Lijie  and
      Wang, Di",
    editor = "Al-Onaizan, Yaser  and
      Bansal, Mohit  and
      Chen, Yun-Nung",
    booktitle = "Proceedings of the 2024 Conference on Empirical Methods in Natural Language Processing",
    month = nov,
    year = "2024",
    address = "Miami, Florida, USA",
    publisher = "Association for Computational Linguistics",
    url = "https://aclanthology.org/2024.emnlp-main.231/",
    doi = "10.18653/v1/2024.emnlp-main.231",
    pages = "3980--3993"
}

@inproceedings{meisenbacher2024dp,
    title = "{DP}-{MLM}: Differentially Private Text Rewriting Using Masked Language Models",
    author = "Meisenbacher, Stephen  and
      Chevli, Maulik  and
      Vladika, Juraj  and
      Matthes, Florian",
    editor = "Ku, Lun-Wei  and
      Martins, Andre  and
      Srikumar, Vivek",
    booktitle = "Findings of the Association for Computational Linguistics: ACL 2024",
    month = aug,
    year = "2024",
    address = "Bangkok, Thailand",
    publisher = "Association for Computational Linguistics",
    url = "https://aclanthology.org/2024.findings-acl.554/",
    doi = "10.18653/v1/2024.findings-acl.554",
    pages = "9314--9328"
}

@inproceedings{macko2024authorship,
    title = "Authorship Obfuscation in Multilingual Machine-Generated Text Detection",
    author = "Macko, Dominik  and
      Moro, Robert  and
      Uchendu, Adaku  and
      Srba, Ivan  and
      Lucas, Jason S  and
      Yamashita, Michiharu  and
      Tripto, Nafis Irtiza  and
      Lee, Dongwon  and
      Simko, Jakub  and
      Bielikova, Maria",
    editor = "Al-Onaizan, Yaser  and
      Bansal, Mohit  and
      Chen, Yun-Nung",
    booktitle = "Findings of the Association for Computational Linguistics: EMNLP 2024",
    month = nov,
    year = "2024",
    address = "Miami, Florida, USA",
    publisher = "Association for Computational Linguistics",
    url = "https://aclanthology.org/2024.findings-emnlp.369/",
    doi = "10.18653/v1/2024.findings-emnlp.369",
    pages = "6348--6368"
}

@inproceedings{xu-etal-2021-utilitarian,
    title = "On a Utilitarian Approach to Privacy Preserving Text Generation",
    author = "Xu, Zekun  and
      Aggarwal, Abhinav  and
      Feyisetan, Oluwaseyi  and
      Teissier, Nathanael",
    editor = "Feyisetan, Oluwaseyi  and
      Ghanavati, Sepideh  and
      Malmasi, Shervin  and
      Thaine, Patricia",
    booktitle = "Proceedings of the Third Workshop on Privacy in Natural Language Processing",
    month = jun,
    year = "2021",
    address = "Online",
    publisher = "Association for Computational Linguistics",
    url = "https://aclanthology.org/2021.privatenlp-1.2/",
    doi = "10.18653/v1/2021.privatenlp-1.2",
    pages = "11--20"
}

@inproceedings{carvalho2023tem,
  title={{TEM}: High utility metric differential privacy on text},
  author={Carvalho, Ricardo Silva and Vasiloudis, Theodore and Feyisetan, Oluwaseyi and Wang, Ke},
  booktitle={Proceedings of the 2023 SIAM International Conference on Data Mining (SDM)},
  pages={883--890},
doi={10.1137/1.9781611977653.ch99},
  year={2023},
  organization={SIAM}
}

@inproceedings{yue-etal-2021-differential,
    title = "Differential Privacy for Text Analytics via Natural Text Sanitization",
    author = "Yue, Xiang  and
      Du, Minxin  and
      Wang, Tianhao  and
      Li, Yaliang  and
      Sun, Huan  and
      Chow, Sherman S. M.",
    editor = "Zong, Chengqing  and
      Xia, Fei  and
      Li, Wenjie  and
      Navigli, Roberto",
    booktitle = "Findings of the Association for Computational Linguistics: ACL-IJCNLP 2021",
    month = aug,
    year = "2021",
    address = "Online",
    publisher = "Association for Computational Linguistics",
    url = "https://aclanthology.org/2021.findings-acl.337/",
    doi = "10.18653/v1/2021.findings-acl.337",
    pages = "3853--3866"
}

@inproceedings{chen-etal-2023-customized,
    title = "A Customized Text Sanitization Mechanism with Differential Privacy",
    author = "Chen, Sai  and
      Mo, Fengran  and
      Wang, Yanhao  and
      Chen, Cen  and
      Nie, Jian-Yun  and
      Wang, Chengyu  and
      Cui, Jamie",
    editor = "Rogers, Anna  and
      Boyd-Graber, Jordan  and
      Okazaki, Naoaki",
    booktitle = "Findings of the Association for Computational Linguistics: ACL 2023",
    month = jul,
    year = "2023",
    address = "Toronto, Canada",
    publisher = "Association for Computational Linguistics",
    url = "https://aclanthology.org/2023.findings-acl.355/",
    doi = "10.18653/v1/2023.findings-acl.355",
    pages = "5747--5758"
}

@inproceedings{maas-etal-2011-learning,
    title = "Learning Word Vectors for Sentiment Analysis",
    author = "Maas, Andrew L.  and
      Daly, Raymond E.  and
      Pham, Peter T.  and
      Huang, Dan  and
      Ng, Andrew Y.  and
      Potts, Christopher",
    editor = "Lin, Dekang  and
      Matsumoto, Yuji  and
      Mihalcea, Rada",
    booktitle = "Proceedings of the 49th Annual Meeting of the Association for Computational Linguistics: Human Language Technologies",
    month = jun,
    year = "2011",
    address = "Portland, Oregon, USA",
    publisher = "Association for Computational Linguistics",
    url = "https://aclanthology.org/P11-1015/",
    pages = "142--150"
}

@inproceedings{
merity2017pointer,
title={Pointer Sentinel Mixture Models},
author={Stephen Merity and Caiming Xiong and James Bradbury and Richard Socher},
booktitle={International Conference on Learning Representations},
year={2017},
url={https://openreview.net/forum?id=Byj72udxe}
}

@inproceedings{10.5555/2969239.2969312,
author = {Zhang, Xiang and Zhao, Junbo and LeCun, Yann},
title = {Character-level convolutional networks for text classification},
year = {2015},
publisher = {MIT Press},
address = {Cambridge, MA, USA},
booktitle = {Proceedings of the 29th International Conference on Neural Information Processing Systems - Volume 1},
pages = {649–657},
numpages = {9},
location = {Montreal, Canada},
series = {NIPS'15},
url = {https://proceedings.neurips.cc/paper_files/paper/2015/file/250cf8b51c773f3f8dc8b4be867a9a02-Paper.pdf}
}

@inproceedings{utpala-etal-2023-locally,
    title = "Locally Differentially Private Document Generation Using Zero Shot Prompting",
    author = "Utpala, Saiteja  and
      Hooker, Sara  and
      Chen, Pin-Yu",
    editor = "Bouamor, Houda  and
      Pino, Juan  and
      Bali, Kalika",
    booktitle = "Findings of the Association for Computational Linguistics: EMNLP 2023",
    month = dec,
    year = "2023",
    address = "Singapore",
    publisher = "Association for Computational Linguistics",
    url = "https://aclanthology.org/2023.findings-emnlp.566/",
    doi = "10.18653/v1/2023.findings-emnlp.566",
    pages = "8442--8457"
}

@article{norgeot2020protected,
  title={Protected Health Information filter (Philter): accurately and securely de-identifying free-text clinical notes},
  author={Norgeot, Beau and Muenzen, Kathleen and Peterson, Thomas A and Fan, Xuancheng and Glicksberg, Benjamin S and Schenk, Gundolf and Rutenberg, Eugenia and Oskotsky, Boris and Sirota, Marina and Yazdany, Jinoos and others},
  journal={NPJ digital medicine},
  volume={3},
  number={1},
  pages={57},
  year={2020},
  publisher={Nature Publishing Group UK London},
  doi={10.1038/s41746-020-0258-y}
}

@misc{kleinberg2022textwashautomatedopensource,
      title={Textwash -- automated open-source text anonymisation}, 
      author={Bennett Kleinberg and Toby Davies and Maximilian Mozes},
      year={2022},
      eprint={2208.13081},
      archivePrefix={arXiv},
      primaryClass={cs.CL},
      url={https://arxiv.org/abs/2208.13081}, 
      doi={10.48550/arXiv.2208.13081}
}

@inproceedings{10.1145/3643651.3659896,
author = {Meisenbacher, Stephen and Chevli, Maulik and Matthes, Florian},
title = {1-Diffractor: Efficient and Utility-Preserving Text Obfuscation Leveraging Word-Level Metric Differential Privacy},
year = {2024},
isbn = {9798400705564},
publisher = {Association for Computing Machinery},
address = {New York, NY, USA},
url = {https://doi.org/10.1145/3643651.3659896},
doi = {10.1145/3643651.3659896},
booktitle = {Proceedings of the 10th ACM International Workshop on Security and Privacy Analytics},
pages = {23–33},
numpages = {11},
location = {Porto, Portugal},
series = {IWSPA '24}
}

@inproceedings{igamberdiev-habernal-2023-dp,
    title = "{DP}-{BART} for Privatized Text Rewriting under Local Differential Privacy",
    author = "Igamberdiev, Timour  and
      Habernal, Ivan",
    editor = "Rogers, Anna  and
      Boyd-Graber, Jordan  and
      Okazaki, Naoaki",
    booktitle = "Findings of the Association for Computational Linguistics: ACL 2023",
    month = jul,
    year = "2023",
    address = "Toronto, Canada",
    publisher = "Association for Computational Linguistics",
    url = "https://aclanthology.org/2023.findings-acl.874/",
    doi = "10.18653/v1/2023.findings-acl.874",
    pages = "13914--13934"
}

@inproceedings{yang-etal-2025-robust,
    title = "Robust Utility-Preserving Text Anonymization Based on Large Language Models",
    author = "Yang, Tianyu  and
      Zhu, Xiaodan  and
      Gurevych, Iryna",
    editor = "Che, Wanxiang  and
      Nabende, Joyce  and
      Shutova, Ekaterina  and
      Pilehvar, Mohammad Taher",
    booktitle = "Proceedings of the 63rd Annual Meeting of the Association for Computational Linguistics (Volume 1: Long Papers)",
    month = jul,
    year = "2025",
    address = "Vienna, Austria",
    publisher = "Association for Computational Linguistics",
    url = "https://aclanthology.org/2025.acl-long.1404/",
    doi = "10.18653/v1/2025.acl-long.1404",
    pages = "28922--28941",
    ISBN = "979-8-89176-251-0"
}

@inproceedings{zaratiana-etal-2024-gliner,
    title = "{GL}i{NER}: Generalist Model for Named Entity Recognition using Bidirectional Transformer",
    author = "Zaratiana, Urchade  and
      Tomeh, Nadi  and
      Holat, Pierre  and
      Charnois, Thierry",
    editor = "Duh, Kevin  and
      Gomez, Helena  and
      Bethard, Steven",
    booktitle = "Proceedings of the 2024 Conference of the North American Chapter of the Association for Computational Linguistics: Human Language Technologies (Volume 1: Long Papers)",
    month = jun,
    year = "2024",
    address = "Mexico City, Mexico",
    publisher = "Association for Computational Linguistics",
    url = "https://aclanthology.org/2024.naacl-long.300/",
    doi = "10.18653/v1/2024.naacl-long.300",
    pages = "5364--5376"
}

@inproceedings{krishna-etal-2021-adept,
    title = "{AD}e{PT}: Auto-encoder based Differentially Private Text Transformation",
    author = "Krishna, Satyapriya  and
      Gupta, Rahul  and
      Dupuy, Christophe",
    editor = "Merlo, Paola  and
      Tiedemann, Jorg  and
      Tsarfaty, Reut",
    booktitle = "Proceedings of the 16th Conference of the European Chapter of the Association for Computational Linguistics: Main Volume",
    month = apr,
    year = "2021",
    address = "Online",
    publisher = "Association for Computational Linguistics",
    url = "https://aclanthology.org/2021.eacl-main.207/",
    doi = "10.18653/v1/2021.eacl-main.207",
    pages = "2435--2439"
}

@inproceedings{jin-etal-2019-pubmedqa,
    title = "{P}ub{M}ed{QA}: A Dataset for Biomedical Research Question Answering",
    author = "Jin, Qiao  and
      Dhingra, Bhuwan  and
      Liu, Zhengping  and
      Cohen, William  and
      Lu, Xinghua",
    editor = "Inui, Kentaro  and
      Jiang, Jing  and
      Ng, Vincent  and
      Wan, Xiaojun",
    booktitle = "Proceedings of the 2019 Conference on Empirical Methods in Natural Language Processing and the 9th International Joint Conference on Natural Language Processing (EMNLP-IJCNLP)",
    month = nov,
    year = "2019",
    address = "Hong Kong, China",
    publisher = "Association for Computational Linguistics",
    url = "https://aclanthology.org/D19-1259/",
    doi = "10.18653/v1/D19-1259",
    pages = "2567--2577"
}

@inproceedings{meisenbacher-etal-2025-leveraging,
    title = "Leveraging Semantic Triples for Private Document Generation with Local Differential Privacy Guarantees",
    author = "Meisenbacher, Stephen  and
      Chevli, Maulik  and
      Matthes, Florian",
    editor = "Christodoulopoulos, Christos  and
      Chakraborty, Tanmoy  and
      Rose, Carolyn  and
      Peng, Violet",
    booktitle = "Proceedings of the 2025 Conference on Empirical Methods in Natural Language Processing",
    month = nov,
    year = "2025",
    address = "Suzhou, China",
    publisher = "Association for Computational Linguistics",
    url = "https://aclanthology.org/2025.emnlp-main.455/",
    doi = "10.18653/v1/2025.emnlp-main.455",
    pages = "8976--8992",
    ISBN = "979-8-89176-332-6"
}

@misc{Hou_Shang_Long_Fu_Chen_2025,
      title={A General Pseudonymization Framework for Cloud-Based LLMs: Replacing Privacy Information in Controlled Text Generation}, 
      author={Shilong Hou and Ruilin Shang and Zi Long and Xianghua Fu and Yin Chen},
      year={2025},
      eprint={2502.15233},
      archivePrefix={arXiv},
      primaryClass={cs.CR},
      url={https://arxiv.org/abs/2502.15233}, 
      doi={10.48550/arXiv.2502.15233}
}

@article{Mishra_Pagare_Sharma_2025, title={A hybrid rule-based NLP and machine learning approach for PII detection and anonymization in financial documents}, volume={15}, rights={2025 The Author(s)}, ISSN={2045-2322}, DOI={10.1038/s41598-025-04971-9}, abstractNote={Safeguarding Personally Identifiable Information (PII) in financial documents is essential to prevent data breaches and maintain regulatory compliance. This research presents a scalable hybrid approach that integrates rule-based Natural Language Processing (NLP), Machine Learning (ML) approaches, and a custom Named Entity Recognition (NER) model for the accurate detection and anonymization of Personally Identifiable Information (PII). A varied and accurate synthetic dataset was created to replicate genuine financial document formats, enhancing model training and assessment. The model has attained a precision of 94.7%, a recall of 89.4%, an F1-score of 91.1%, and an overall accuracy of 89.4% on synthetic datasets. Additional validation on actual financial documents, such as audit reports and vendor bills, revealed a consistent performance with an accuracy of 93%. The study utilizes confusion matrices, ROC curves, and precision-recall curves to evaluate the model which further validates the model’s capabilities and generalization ability. The suggested approach provides a robust and efficient solution for protecting sensitive information in operational financial contexts, markedly enhancing current methods for PII protection.}, number={1}, journal={Scientific Reports}, publisher={Nature Publishing Group}, author={Mishra, Kushagra and Pagare, Harsh and Sharma, Kanhaiya}, year={2025}, month={7}, pages={22729}, language={en} }

@inproceedings{Meisenbacher_Kleinert_Matthes_2026,
    title = "A Systematic Exploration of Text Decomposition and Budget Distribution in Differentially Private Text Obfuscation",
    author = "Meisenbacher, Stephen  and
      Kleinert, Angelo  and
      Matthes, Florian",
    editor = "Habernal, Ivan  and
      Ghanavati, Sepideh  and
      Haghighi, Sara  and
      Ramesh, Krithika  and
      Igamberdiev, Timour  and
      Wilson, Shomir",
    booktitle = "Proceedings of the Seventh Workshop on Privacy in Natural Language Processing",
    month = jul,
    year = "2026",
    address = "San Diego, California",
    publisher = "Association for Computational Linguistics",
    url = "https://aclanthology.org/2026.privatenlp-main.9/",
    doi = "10.18653/v1/2026.privatenlp-main.9",
    pages = "118--139",
    ISBN = "979-8-89176-397-5"
}

@article{isaoglu2026unified,
  title={A Unified Evaluation Framework for Utility and Privacy Risks of LLM-Generated Synthetic Text Data},
  author={Isaoglu, Lubana and Orman, Zeynep},
  journal={Engineering Perspective},
  volume={6},
  number={3},
  pages={456--466},
  year={2026},
  publisher={Hamit Solmaz},
  doi={10.64808/engineeringperspective.1910777}
}

@inproceedings{
Hu_McKenna_2026,
title={{ACTG}-{ARL}: Differentially Private Conditional Text Generation with {RL}-Boosted Control},
author={Yuzheng Hu and Ryan McKenna and Da Yu and Shanshan Wu and Han Zhao and Zheng Xu and Peter Kairouz},
booktitle={Forty-third International Conference on Machine Learning},
year={2026},
url={https://openreview.net/forum?id=vvke6mnh3Z}
}

@inproceedings{zhao-field-2025-controlled,
    title = "Controlled Generation for Private Synthetic Text",
    author = "Zhao, Zihao  and
      Field, Anjalie",
    editor = "Christodoulopoulos, Christos  and
      Chakraborty, Tanmoy  and
      Rose, Carolyn  and
      Peng, Violet",
    booktitle = "Proceedings of the 2025 Conference on Empirical Methods in Natural Language Processing",
    month = nov,
    year = "2025",
    address = "Suzhou, China",
    publisher = "Association for Computational Linguistics",
    url = "https://aclanthology.org/2025.emnlp-main.1663/",
    doi = "10.18653/v1/2025.emnlp-main.1663",
    pages = "32720--32735",
    ISBN = "979-8-89176-332-6"
}

@inproceedings{Faveri_Faggioli_Ferro_2025, address={New York, NY, USA}, series={CIKM ’25}, title={DP-COMET: A Differential Privacy Contextual Obfuscation MEchanism for Texts in Natural Language Processing}, ISBN={979-8-4007-2040-6}, url={https://dl.acm.org/doi/10.1145/3746252.3760888}, DOI={10.1145/3746252.3760888}, abstractNote={Protecting sensitive information within textual data strongly depends on the context in which the data is presented. However, current privacy-preserving obfuscation mechanisms based on epsilon-Differential Privacy (DP) produce an obfuscated private text, changing the original phrase term-by-term without considering the context in which such a term is placed. This paper introduces DP-COMET, an epsilon-DP obfuscation mechanism that evaluates a text’s context before producing its private version. The mechanism defines a representation of the original text that considers the entire context within the text, producing an obfuscated version after adding noise to this representation and depending on the privacy parameter epsilon. We test DP-COMET on different Natural Language Processing (NLP) and Information Retrieval (IR) downstream tasks, and our findings show that our obfuscation mechanism not only achieves comparable performance results to traditional term-by-term mechanisms but also produces obfuscated texts less similar to the originals. To promote the reproducibility of DP-COMET, we make the code publicly available at https://github.com/Kekkodf/DP-COMET.}, booktitle={Proceedings of the 34th ACM International Conference on Information and Knowledge Management}, publisher={Association for Computing Machinery}, author={De Faveri, Francesco Luigi and Faggioli, Guglielmo and Ferro, Nicola}, year={2025}, month={11}, pages={4700–4705}, collection={CIKM ’25} }

@inproceedings{Li_Fan_Fu_Ding_Feng,
    title = "{DP}-{GTR}: Differentially Private Prompt Protection via Group Text Rewriting",
    author = "Li, Mingchen  and
      Fan, Heng  and
      Fu, Song  and
      Ding, Junhua  and
      Feng, Yunhe",
    editor = "Christodoulopoulos, Christos  and
      Chakraborty, Tanmoy  and
      Rose, Carolyn  and
      Peng, Violet",
    booktitle = "Findings of the Association for Computational Linguistics: EMNLP 2025",
    month = nov,
    year = "2025",
    address = "Suzhou, China",
    publisher = "Association for Computational Linguistics",
    url = "https://aclanthology.org/2025.findings-emnlp.83/",
    doi = "10.18653/v1/2025.findings-emnlp.83",
    pages = "1573--1585",
    ISBN = "979-8-89176-335-7"
}

@inproceedings{Sun_Tian_Song_2025, address={Vienna, Austria}, title={DPGA-TextSyn: Differentially Private Genetic Algorithm for Synthetic Text Generation}, ISBN={979-8-89176-256-5}, url={https://aclanthology.org/2025.findings-acl.831/}, DOI={10.18653/v1/2025.findings-acl.831}, abstractNote={Using large language models (LLMs) has a potential risk of privacy leakage since the data with sensitive information may be used for fine-tuning the LLMs. Differential privacy (DP) provides theoretical guarantees of privacy protection, but its practical application in LLMs still has the problem of privacy-utility trade-off. Researchers synthesized data with strong generation capabilities closed-source LLMs (i.e., GPT-4) under DP to alleviate this problem, but this method is not so flexible in fitting the given privacy distributions without fine-tuning. Besides, such methods can hardly balance the diversity of synthetic data and its relevance to target privacy data without accessing so much private data. To this end, this paper proposes DPGA-TextSyn, combining general LLMs with genetic algorithm (GA) to produce relevant and diverse synthetic text under DP constraints. First, we integrate the privacy gene (i.e., metadata) to generate better initial samples. Then, to achieve survival of the fittest and avoid homogeneity, we use privacy nearest neighbor voting and similarity suppression to select elite samples. In addition, we expand elite samples via genetic strategies such as mutation, crossover, and generation to expand the search scope of GA. Experiments show that this method significantly improves the performance of the model in downstream tasks while ensuring privacy.}, booktitle={Findings of the Association for Computational Linguistics: ACL 2025}, publisher={Association for Computational Linguistics}, author={Sun, Zhonghao and Tian, Zhiliang and Song, Yiping and Si, Yuyi and Zhang, Juhua and Huang, Minlie and Lu, Kai and Xiong, Zeyu and Liu, Xinwang and Li, Dongsheng}, editor={Che, Wanxiang and Nabende, Joyce and Shutova, Ekaterina and Pilehvar, Mohammad Taher}, year={2025}, month={7}, pages={16159–16179} }

@inproceedings{Sun_Guo_2026, title={DPMasker: A Context-Aware Reversible Differential Privacy Perturbation Algorithm for Medical Text}, url={https://ieeexplore.ieee.org/abstract/document/11508739}, DOI={10.1109/ICAACE69793.2026.11508739}, abstractNote={Automatic clinical text anonymization is crucial for unlocking the secondary application of Electronic Health Records (EHRs) in medical research, yet existing methods often struggle to strike a balance among privacy protection strength, semantic preservation, and data reversibility. Traditional methods based on Named Entity Recognition (NER) and rewriting are prone to poor generalization or the disruption of medical contexts, while differential privacy (DP) perturbation based on vector distance is often constrained by rigid candidate word distributions. To address these issues, this paper proposes DPMasker, a reversible privacy perturbation algorithm for medical texts. Rather than relying on vector distances, this algorithm utilizes Bio-ClinicalBERT to extract the contextual feature distribution of target entities. By introducing a dual-stage dynamic truncation mechanism (head risk isolation and nucleus truncation semantic filtering), DPMasker strictly removes high-risk privacy words and long-tail noise that disrupts medical logic. Subsequently, controlled sampling is performed on a secure candidate set combining the DP exponential mechanism, achieving mathematically strict, quantifiable privacy constraints and maximizing semantic utility. Furthermore, the algorithm designs a DP embedding hash mapping storage mechanism based on HMACSHA256 to meet the zero-trust secure traceback requirements of high-privilege hospital environments. Extensive experiments on the HealthCareMagic real-world doctor-patient dialogue dataset demonstrate that DPMasker successfully achieves absolute zero leakage against privacy extraction attacks. Meanwhile, in the generative utility evaluation of large language models (LLMs) such as Qwen3, GLM-4.6, and Kimi-K2, its BLEU and ROUGE-L scores significantly surpass those of existing mainstream baseline methods.}, booktitle={2026 9th International Conference on Advanced Algorithms and Control Engineering (ICAACE)}, author={Sun, Yifan and Guo, Weibin}, year={2026}, month={3}, pages={2374–2380} }

@inproceedings{Zhang_Tian_2025, address={Vienna, Austria}, title={DYNTEXT: Semantic-Aware Dynamic Text Sanitization for Privacy-Preserving LLM Inference}, ISBN={979-8-89176-256-5}, url={https://aclanthology.org/2025.findings-acl.1038/}, DOI={10.18653/v1/2025.findings-acl.1038}, abstractNote={LLMs face privacy risks when handling sensitive data. To ensure privacy, researchers use differential privacy (DP) to provide protection by adding noise during LLM training. However, users may be hesitant to share complete data with LLMs. Researchers follow local DP to sanitize the text on the user side and feed non-sensitive text to LLMs. The sanitization usually uses a fixed non-sensitive token list or a fixed noise distribution, which induces the risk of being attacked or semantic distortion. We argue that the token’s protection level should be adaptively adjusted according to its semantic-based information to balance the privacy-utility trade-off. In this paper, we propose DYNTEXT, an LDP-based Dynamic Text sanitization for privacy-preserving LLM inference, which dynamically constructs semantic-aware adjacency lists of sensitive tokens to sample non-sensitive tokens for perturbation. Specifically, DYNTEXT first develops a semantic-based density modeling under DP to extract each token’s density information. We propose token-level smoothing sensitivity by combining the idea of global sensitivity (GS) and local sensitivity (LS), which dynamically adjusts the noise scale to avoid excessive noise in GS and privacy leakage in LS. Then, we dynamically construct an adjacency list for each sensitive token based on its semantic density information. Finally, we apply the replacement mechanism to sample non-sensitive, semantically similar tokens from the adjacency list to replace sensitive tokens. Experiments show that DYNTEXT excels strong baselines on three datasets.}, booktitle={Findings of the Association for Computational Linguistics: ACL 2025}, publisher={Association for Computational Linguistics}, author={Zhang, Juhua and Tian, Zhiliang and Zhu, Minghang and Song, Yiping and Sheng, Taishu and Yang, Siyi and Du, Qiunan and Liu, Xinwang and Huang, Minlie and Li, Dongsheng}, editor={Che, Wanxiang and Nabende, Joyce and Shutova, Ekaterina and Pilehvar, Mohammad Taher}, year={2025}, month={7}, pages={20243–20255} }

@inproceedings{Sun_Schlegel_2025, address={New York, NY, USA}, series={CIKM ’25}, title={Evaluating Differentially Private Generation of Domain-Specific Text}, ISBN={979-8-4007-2040-6}, url={https://dl.acm.org/doi/10.1145/3746252.3760916}, DOI={10.1145/3746252.3760916}, abstractNote={Generative AI offers transformative potential for high-stakes domains such as healthcare and finance, yet privacy and regulatory barriers hinder the use of real-world data. To address this, differentially private synthetic data generation has emerged as a promising alternative. In this work, we introduce a unified benchmark to systematically evaluate the utility and fidelity of text datasets generated under formal Differential Privacy (DP) guarantees. Our benchmark addresses key challenges in domain-specific benchmarking, including choice of representative data and realistic privacy budgets, accounting for pre-training and a variety of evaluation metrics. We assess state-of-the-art privacy-preserving generation methods across five domain-specific datasets, revealing significant utility and fidelity degradation compared to real data, especially under strict privacy constraints. These findings underscore the limitations of current approaches, outline the need for advanced privacy-preserving data sharing methods and set a precedent regarding their evaluation in realistic scenarios.}, booktitle={Proceedings of the 34th ACM International Conference on Information and Knowledge Management}, publisher={Association for Computing Machinery}, author={Sun, Yidan and Schlegel, Viktor and Kolumam Nandakumar, Srinivasan and Zahid, Iqra and Wu, Yuping and Del-Pinto, Warren and Nenadic, Goran and Siew Kei, Lam and Zhang, Jie and Bharath, Anil}, year={2025}, month={11}, pages={5273–5278}, collection={CIKM ’25} }

@inproceedings{Karousos_Vorvilas_Pantazi_Verykios_2025, title={Evaluating NER Approaches to Support Qualitative Educational Data Anonymization in the Greek Language: A Comparative Study}, url={https://ieeexplore.ieee.org/abstract/document/11311218}, DOI={10.1109/IISA66859.2025.11311218}, abstractNote={The present research focus on the use of Named Entity Recognition (NER) for the anonymization of qualitative educational data in Greek. More specifically, it investigates the effectiveness of NER models for the Greek language in identifying sensitive information, such as tutor names, within open-ended student responses collected from course evaluation surveys at the Hellenic Open University (HOU). Five different NER models were examined, both individually and in combinations, to determine their performance in identifying proper names. The findings show that while all models demonstrate high precision and recall for non-entity classes, however significant differences are found in their ability to identify named entities. The Toolkit model is the best performing model, achieving high recall and precision, with some combinations of NER models improving recall but resulting in lower precision. The findings underscore the criticality of choosing appropriate NER models, especially in cases that involve category imbalance, and presents several trade-offs between precision and recall that one should consider in NER and anonymization tasks.}, booktitle={2025 16th International Conference on Information, Intelligence, Systems \& Applications (IISA)}, author={Karousos, Nikos and Vorvilas, George and Pantazi, Despoina and Verykios, Vassilios S.}, year={2025}, month={7}, pages={1–8} }

@inproceedings{Frikha_Walha_2025, address={Mumbai, India}, title={IncogniText: Privacy-enhancing Conditional Text Anonymization via LLM-based Private Attribute Randomization}, ISBN={979-8-89176-298-5}, url={https://aclanthology.org/2025.ijcnlp-long.134/}, DOI={10.18653/v1/2025.ijcnlp-long.134}, abstractNote={In this work, we address the problem of text anonymization where the goal is to prevent adversaries from correctly inferring private attributes of the author, while keeping the text utility, i.e., meaning and semantics. We propose IncogniText, a technique that anonymizes the text to mislead a potential adversary into predicting a wrong private attribute value. Our empirical evaluation shows a reduction of private attribute leakage by more than across 8 different private attributes. Finally, we demonstrate the maturity of IncogniText for real-world applications by distilling its anonymization capability into a set of LoRA parameters associated with an on-device model. Our results show the possibility of reducing privacy leakage by more than half with limited impact on utility.}, booktitle={Proceedings of the 14th International Joint Conference on Natural Language Processing and the 4th Conference of the Asia-Pacific Chapter of the Association for Computational Linguistics}, publisher={The Asian Federation of Natural Language Processing and The Association for Computational Linguistics}, author={Frikha, Ahmed and Walha, Nassim and Nakka, Krishna Kanth and Mendes, Ricardo and Jiang, Xue and Zhou, Xuebing}, editor={Inui, Kentaro and Sakti, Sakriani and Wang, Haofen and Wong, Derek F. and Bhattacharyya, Pushpak and Banerjee, Biplab and Ekbal, Asif and Chakraborty, Tanmoy and Singh, Dhirendra Pratap}, year={2025}, month={12}, pages={2490–2501} }

@article{Karamitsos_Roufas_2025, title={LegNER: a domain-adapted transformer for legal named entity recognition and text anonymization}, volume={8}, ISSN={2624-8212}, url={https://www.frontiersin.org/journals/artificial-intelligence/articles/10.3389/frai.2025.1638971/full}, DOI={10.3389/frai.2025.1638971}, abstractNote={The increasing demand for scalable and privacy-preserving processing of legal documents has intensified the need for accurate Named Entity Recognition (NER) systems tailored to the legal domain. In this work, we introduce LegNER, a domain-adapted transformer model designed for both legal NER and text anonymization. The model is trained on a corpus of 1,542 manually annotated court cases and enriched with an extended legal vocabulary, enabling robust recognition of six critical entity types, including PERSON, ORGANIZATION, LAW, and CASE_REFERENCE. Built on BERT-base and enhanced through domain-specific pretraining and span-level supervision, LegNER consistently outperforms established legal NER baselines. Experimental results demonstrate significant gains in accuracy (99%), F1 score (over 99%), and inference efficiency (processing more than 12 documents per second), confirming both its precision and scalability. Beyond quantitative improvements, qualitative evaluation highlights LegNERs ability to generate coherent anonymized outputs, a crucial requirement for GDPR-compliant redaction and automated legal analytics. Taken together, these results establish LegNER as a reliable and effective solution for high-precision entity recognition and anonymization in compliance-sensitive legal workflows.}, journal={Frontiers in Artificial Intelligence}, publisher={Frontiers}, author={Karamitsos, Ioannis and Roufas, Nikolaos and Al-Hussaeni, Khalil and Kanavos, Andreas}, year={2025}, month={11}, language={English} }

@article{Papadopoulou_2023,
  title={Neural text sanitization with privacy risk indicators: An empirical analysis},
  author={Papadopoulou, Anthi and Lison, Pierre and Anderson, Mark and {\O}vrelid, Lilja and Pil{\'a}n, Ildik{\'o}},
  journal={Language Resources and Evaluation},
  volume={60},
  number={2},
  pages={32},
  year={2026},
  publisher={Springer},
  doi={10.1007/s10579-026-09911-1}
}

@inproceedings{Mancera_Morales_2026,
  title={{PBa-LLM}: Privacy-and bias-aware NLP using named-entity recognition (NER)},
  author={Mancera, Gonzalo and Morales, Aythami and Fierrez, Julian and Tolosana, Ruben and Pe{\~n}a, Alejandro and Lopez-Duran, Miguel and Jurado, Francisco and Ortigosa, Alvaro},
  booktitle={International Conference on Document Analysis and Recognition},
  pages={3--20},
  year={2025},
  organization={Springer},
  doi={10.1007/978-3-032-09368-4_1}
}

@article{kim2025privrewrite,
  title={PrivRewrite: Differentially Private Text Rewriting Under Black-Box Access with Refined Sensitivity Guarantees},
  author={Kim, Jongwook},
  journal={Applied Sciences},
  volume={15},
  number={22},
  pages={11930},
  year={2025},
  publisher={MDPI},
  doi={10.3390/app152211930}
}

@book{bernard2016analyzing,
  title={Analyzing qualitative data: Systematic approaches},
  author={Bernard, H Russell and Wutich, Amber and Ryan, Gery W},
  year={2016},
  publisher={SAGE publications}
}

\newpage
\appendix
\onecolumn
\section{Complete Literature Review Analysis Results}

\begin{table}[h!]
\centering
\small
\setlength{\tabcolsep}{4pt}
\resizebox*{!}{0.92\textheight}{
\begin{tabular}{p{0.10\textwidth} | p{0.2\textwidth} | c | p{1.1\textwidth}}
\hline
\textbf{Category} & \textbf{Evaluation Method} & \textbf{\#} & \textbf{Supporting Papers} \\
\hline
\multirow{18}{*}{\parbox{0.12\textwidth}{\raggedright Empirical privacy \& attacks}}
& Attack success rate (ASR) & 28 &  \cite{macko2024authorship, tong2025inferdpt, li2024llm, zeng2025privacyrestore, vats2024recovering, meisenbacher2024spend, uchendu2023attribution, chen2024combating, singh2024enhancing, aditya2024evaluating, feyisetan2019leveraging, patsakis2023man, shao2024quantifying, chow2009sanitization, du2023sanitizing, mai2023split, khadam2020text, gu2024towards, fernandes2019generalised, khadam2021advanced, mimoto2020practical, gardiner2024data, larbi2023clinical, ren-etal-2025-measure, Meisenbacher_Chevli_Matthes_2025a, Tong_Chen_Yuan_2025, Kuo_Zhang_Zhang_Tang_2025, yang-etal-2025-robust} \\
\cline{2-4}
& (Carlini) exposure (rate) & 12 & \cite{sun2024effectiveness, li2024fine, li-etal-2024-privlm, shi2022selective, shi2022just, anil2022large, hoory2021learning, smith2024identifying, wu2023depn, sousa2023keep, gu2024towards, Zhang_2025} \\
\cline{2-4}
& Plausible deniability: $N_w$ (prob.\ word returned unperturbed) & 8 & \cite{10.1145/3643651.3659896, huang2024truncatedlaplacian, feyisetan2019leveraging, feyisetan2019perturbations, xu2020differentially, yue-etal-2021-differential, arnold2023driving, arnold2023guiding} \\
\cline{2-4}
& Plausible deniability: $S_w$ (support of perturbing a word) & 6 & \cite{10.1145/3643651.3659896, feyisetan2019leveraging, feyisetan2019perturbations, xu2020differentially, yue-etal-2021-differential, arnold2023guiding} \\
\cline{2-4}
& PII count / percentage & 7 & \cite{vinod2025invisibleink, Mancera_Morales_2026, vats2024recovering, smith2024identifying, qu2021natural, Meisenbacher_Kleinert_Matthes_2026, Zhao_Field_2025} \\
\cline{2-4}
& De-anonymization success / PII reconstruction rate & 8 & \cite{zarski2025enhancing, wang2024knowledgesg, zhou2023textobfuscator, zhao-field-2025-controlled, Sun_Guo_2026, Kalari_Padidela_2026, Pilan_Manzanares_2025, Manzanares-Salor_2026} \\
\cline{2-4}
& Membership / attack advantage (TPR -- FPR) & 3 & \cite{vu2024granularity, wunderlich2022privacy, smith2024identifying} \\
\hline
\multirow{28}{*}{\parbox{0.12\textwidth}{\raggedright Utility}}
& Accuracy / \newline Precision / \newline Recall / \newline F1 & 147 & \cite{10.1145/3643651.3659896,liu2024balancing,salemi2025comparing,yu2021differentially,sun2024effectiveness,zarski2025enhancing,li2024fine,higashi2025differential,wang2024knowledgesg,li2024llm,Mancera_Morales_2026,de2024pypantera,staufer2024silencing,loiseau-etal-2025-tau,plant2025you,kamoji2024privacy,uchendu2023attribution,chen2024combating,singh2024enhancing,Zhang2023EnhancingPI,aditya2024evaluating,resende2022fast,hessel2021effective,smith2024identifying,hao2022iron,feyisetan2019leveraging,patsakis2023man,li2023mope,zhan2021multi,senge2022one,yoon2022optimal,kassem2023preserving,bannihatti2023privacy,wunderlich2022privacy,shao2024quantifying,du2023sanitizing,shree2020sensitivity,mishra2024sentinellms,prabhumoye2018style,carvalho2023tem,tokpo2022text,zhou2023textobfuscator,yu2024textual,chen2022x,li2018towards,ponomareva2022training,lai2022user,hassan2021utility,yu2017vector,beigi2019privacy,shi2022selective,tang2023privacy,wu2023privacy,duan2023flocks,xie2024differentially,mattern2022differentially,qu2021natural,anil2022large,plant2021cape,canfora2018nlp,yue-etal-2021-differential,bo2021er,maheshwari2022fair,sousa2023keep,ye2024openfedllm,liu2024PPTIF,huang2024privacy,matzken2023trade,gu2024towards,larbi2023clinical,nagy2023privacy,yun2023privacy,wiest2024privacy,yin2022privacy,pepin2024privacy,moghaddam2024privacy,song2024private,feyisetan2021private,badawi2020privFT,feyisetan2019perturbations,adelani2020privacy,macko2024authorship,asimopoulos2024benchmarking,tchouka2024differentially,meisenbacher2024dp,meisenbacher2024just,zeng2025privacyrestore,li-etal-2024-privlm,meisenbacher2024spend,raffetseder2025support,meisenbacher2024thinking,hathurusinghe2021privacy,utpala-etal-2023-locally,panov2022mucaat,iwendi2020n,adelani2021preventing,yermilov2023privacy,alawad2020privacy,igamberdiev2022privacy,bannour2022privacy,Luo2024SecFormerTF,meehan2022sentence,weggenmann2018syntf,sanchez2017toward,hoory2021learning,meisenbacher2024collocation,juez2023agora,biesner2022anonymization,sanchez2016c,igamberdiev2023dpbart,weggenmann2022dp,xie2023fused,zecevic2024generation,mcmurry2013improved,gardiner2024data,mouhammad2023crowdsourcing,bao2024keep,carranza2024synthetic,sanchez2014utility, Hou_Shang_Long_Fu_Chen_2025, Mishra_Pagare_Sharma_2025, Meisenbacher_Kleinert_Matthes_2026, isaoglu2026unified, Hu_McKenna_2026, zhao-field-2025-controlled, Wiest_Wermke_Kather_2025, Li_Fan_Fu_Ding_Feng, Sun_Tian_Song_2025, Karousos_Vorvilas_Pantazi_Verykios_2025, ren-etal-2025-measure, Frikha_Walha_2025, Shahriar_Dara_Zarrinkalam_2025, Madaan_Ramesh_2026, Papadopoulou_2023, Psarra_Stefanidis_2026, Meisenbacher_Chevli_Matthes_2025a, Mancera_Morales_2026, Shokri_Levitan_2025, Tobia_Patarnello_2025, Ochs_Habernal_2025, Shi_Yuan_2025, yang-etal-2025-robust, Tian_Bhattacharjee_2026, Meisenbacher_Klymenko_Bodea_Matthes_2025, Manzanares-Salor_2026, Huang_Yuan_Haffari_Qu, Abbasalizadeh_Narain_2025} \\
\cline{2-4}
& AUC-ROC & 15 & \cite{macko2024authorship, li-etal-2024-privlm, smith2024identifying, li2024llm, li2023mope, wunderlich2022privacy, feyisetan2019perturbations, nagy2023privacy, feyisetan2021private, carvalho2023tem, tang2023privacy, duan2023flocks, duan2024privacy, Mishra_Pagare_Sharma_2025, isaoglu2026unified} \\
\cline{2-4}
& Exact match (task EM) & 9 & \cite{zeng2025privacyrestore, khoje2024navigating, alabdulkareem2024securellm, ponomareva2022training, smith2024identifying, Hou_Shang_Long_Fu_Chen_2025, Kuo_Zhang_Zhang_Tang_2025, Shi_Yuan_2025, Tian_Bhattacharjee_2026} \\
\cline{2-4}
& True Positive Rate (TPR) & 6 & \cite{higashi2025differential, li-etal-2024-privlm, smith2024identifying, li2023mope, maheshwari2022fair, li2024llm} \\
\cline{2-4}
& False Positive Rate (FPR) & 7 & \cite{higashi2025differential, li-etal-2024-privlm, resende2022fast,  smith2024identifying, li2023mope, zhao-field-2025-controlled, Wiest_Wermke_Kather_2025} \\
\cline{2-4}
& MCC (Matthews Correlation Coefficient) & 5 & \cite{hessel2021effective, song2024private, Luo2024SecFormerTF, mattern-etal-2022-limits, matzken2023trade} \\
\cline{2-4}
& MRR (Mean Reciprocal Rank) & 4 & \cite{li2024fine, bao2024keep, wu2023depn, feyisetan2019perturbations} \\
\cline{2-4}
& Confusion matrices & 5 & \cite{senge2022one, wiest2024privacy, pepin2024privacy, Mishra_Pagare_Sharma_2025, Karousos_Vorvilas_Pantazi_Verykios_2025} \\
\cline{2-4}
& Utility preservation (\%) from information content (IC) & 3 & \cite{hassan2021utility, sanchez2016c, sanchez2014utility} \\
\hline
\multirow{18}{*}{\parbox{0.12\textwidth}{\raggedright Text quality \newline \& semantic preservation}}
& Perplexity / Generation perplexity gap & 37 & \cite{awon2025clusant, flemings2024differentially_private_knowledge, flemings2024differentially_private_next_token, vinod2025invisibleink, higashi2025differential, vats2024recovering, meisenbacher2024spend, meisenbacher2024thinking, turan2022adapting, li2023beyond, wu2023depn, kerrigan2020differentially, aditya2024evaluating, liu2024forgetting, smith2024identifying, kassem2023preserving, tokpo2022text, mattern-etal-2022-limits, matzken2023trade, lai2022user, shi2022selective, majmudar2022differentially, ginart2022submix, tian2022seqpate, weggenmann2022dp, li2024llm, Meisenbacher_Kleinert_Matthes_2026, isaoglu2026unified, Zhao_Field_2025, Madaan_Ramesh_2026, Meisenbacher_Chevli_Matthes_2025a, Ochs_Habernal_2025, Kuo_Zhang_Zhang_Tang_2025, Huang_Yuan_Haffari_Qu} \\
\cline{2-4}
& Cosine similarity (embedding-based) & 28 & \cite{awon2025clusant, meisenbacher2024dp, zhu2024exploiting, tong2025inferdpt, de2024pypantera, staufer2024silencing, meisenbacher2024spend, meisenbacher2024thinking, uchendu2023attribution, tokpo2022text, mattern-etal-2022-limits, mattern2022differentially, sasada2021differentially, weggenmann2022dp, bo2021er, meisenbacher2024just, jiang2009t, gu2024towards, Meisenbacher_Kleinert_Matthes_2026, zhao-field-2025-controlled, Faveri_Faggioli_Ferro_2025, Sun_Schlegel_2025, meisenbacher-etal-2025-leveraging, Meisenbacher_Chevli_Matthes_2025a, kim2025privrewrite, Tian_Bhattacharjee_2026, Meisenbacher_Klymenko_Bodea_Matthes_2025, Pilan_Manzanares_2025} \\
\cline{2-4}
& ROUGE-1 / ROUGE-2 / ROUGE-L & 25 & \cite{salemi2025comparing, yu2021differentially, hou2025fine, liang2023mergefastprivatetext, huang2025nap2, zeng2025privacyrestore, huang2024truncatedlaplacian, staufer2024silencing, li2023beyond, smith2024identifying, yermilov2023privacy, zhou2023textobfuscator, yu2024textual, zecevic2024generation, gu2024towards, gardiner2024data, wu2023privacy, Hou_Shang_Long_Fu_Chen_2025, Zhao_Field_2025, Li_Fan_Fu_Ding_Feng, Sun_Guo_2026, Frikha_Walha_2025, Madaan_Ramesh_2026, Meisenbacher_Chevli_Matthes_2025a, Huang_Yuan_Haffari_Qu} \\
\cline{2-4}
& BLEU & 21 & \cite{yu2021differentially, higashi2025differential, huang2024truncatedlaplacian, meisenbacher2024spend, carranza2024synthetic, meisenbacher2024thinking, smith2024identifying, feng2020securenlp, prabhumoye2018style, igamberdiev2023dpbart, igamberdiev2024dp_nmt, weggenmann2022dp, ye2024openfedllm, gu2024towards, Hou_Shang_Long_Fu_Chen_2025, Li_Fan_Fu_Ding_Feng, Sun_Guo_2026, Sun_Schlegel_2025, Shao_Liu_2026, Huang_Yuan_Haffari_Qu} \\
\cline{2-4}
& MAUVE  & 12 & \cite{tong2025inferdpt, vinod2025invisibleink, mattern2022differentially, wang2024knowledgesg, carranza2024synthetic, xie2024differentially, uchendu2023attribution, isaoglu2026unified, Hu_McKenna_2026, Zhao_Field_2025, Sun_Tian_Song_2025, Zhang_Tian_2025} \\
\cline{2-4}
& METEOR & 7 & \cite{yu2021differentially, higashi2025differential, fisher2024jamdec, liang2023mergefastprivatetext, uchendu2023attribution, adelani2021preventing, Sun_Schlegel_2025} \\
\cline{2-4}
& BERTScore / BARTScore & 7 & \cite{liang2023mergefastprivatetext, huang2024truncatedlaplacian, utpala-etal-2023-locally,  isaoglu2026unified, Sun_Schlegel_2025, kim2025privrewrite, Shi_Yuan_2025} \\
\cline{2-4}
& Word Error Rate (WER) & 3 & \cite{vats2024recovering, turan2022adapting, liu2024forgetting} \\
\hline
\multirow{6}{*}{\parbox{0.12\textwidth}{\raggedright Regression / error \& correlation}}
& Pearson / Spearman $\rho$ / Correlation & 8 & \cite{hessel2021effective, feyisetan2019leveraging, li2023mope, song2024private, Luo2024SecFormerTF, mishra2024sentinellms, chen2022x, matzken2023trade} \\
\cline{2-4}
& MAE (Mean Absolute Error) & 3 & \cite{salemi2025comparing, liu2024PPTIF, gu2024towards} \\
\cline{2-4}
& Signal to noise ratio (PSNR) & 3 & \cite{yin2022privacy, khadam2021advanced, khadam2020text} \\
\hline
\multirow{12}{*}{\parbox{0.12\textwidth}{\raggedright System / efficiency}}
& Runtime (ms) / Runtime volume & 13 & \cite{resende2022fast, hao2022iron, senge2022one, vatsalan2021topicmodelling, yin2022privacy, sanchez2017toward, yu2017vector, morabito2023docflow, liu2024PPTIF, huang2024privacy, gu2024towards, sanchez2014utility, chakaravarthy2008efficient} \\
\cline{2-4}
& Communication volume / overhead / message sizes & 7 & \cite{hao2022iron, liang2023mergefastprivatetext, Luo2024SecFormerTF, feng2020securenlp, liu2024PPTIF, huang2024privacy, zhao-field-2025-controlled} \\
\cline{2-4}
& Speed & 6 & \cite{liang2023mergefastprivatetext, song2024private, badawi2020privFT, liu2024PPTIF, huang2024privacy, Karamitsos_Roufas_2025} \\
\cline{2-4}
& Total inference time / Per-operation time & 4 & \cite{liang2023mergefastprivatetext, song2024private, Luo2024SecFormerTF, gu2024towards} \\
\cline{2-4}
& (Comp.\ / Comm. / Encrypt./Decrypt.) Latency & 3 &  \cite{Zhang2023EnhancingPI, badawi2020privFT, huang2024privacy} \\
\cline{2-4}
& Memory/disk footprint / RAM usage (engineering metrics) & 3 & \cite{bannihatti2023privacy, vatsalan2021topicmodelling, moghaddam2024privacy} \\
\hline
\multirow{5}{*}{\parbox{0.12\textwidth}{\raggedright Others}}
& LLM score / LLM-Judge (undefined rubric varies across papers) & 9 & \cite{de2025comparative, awon2025clusant, wang2024knowledgesg, huang2025nap2, zeng2025privacyrestore, matzken2023trade, Frikha_Walha_2025, Kuo_Zhang_Zhang_Tang_2025, Huang_Yuan_Haffari_Qu} \\
\cline{2-4}
& Relative gain & 8 & \cite{meisenbacher2024dp, meisenbacher2024thinking, weggenmann2018syntf, mattern-etal-2022-limits, meisenbacher2024collocation, Meisenbacher_Kleinert_Matthes_2026, meisenbacher-etal-2025-leveraging, Meisenbacher_Chevli_Matthes_2025a} \\

\end{tabular}
}
\caption{A summary of the literature review analysis, grouped by (categories of) text privatization evaluation metrics.}
\label{tab:metrics_counts}
\end{table}

\newpage

\section{Detailed System Architecture of PrivBench}
\label{sec:arch}

\begin{figure*}[ht]
    \centering
    \includegraphics[width=0.95\linewidth]{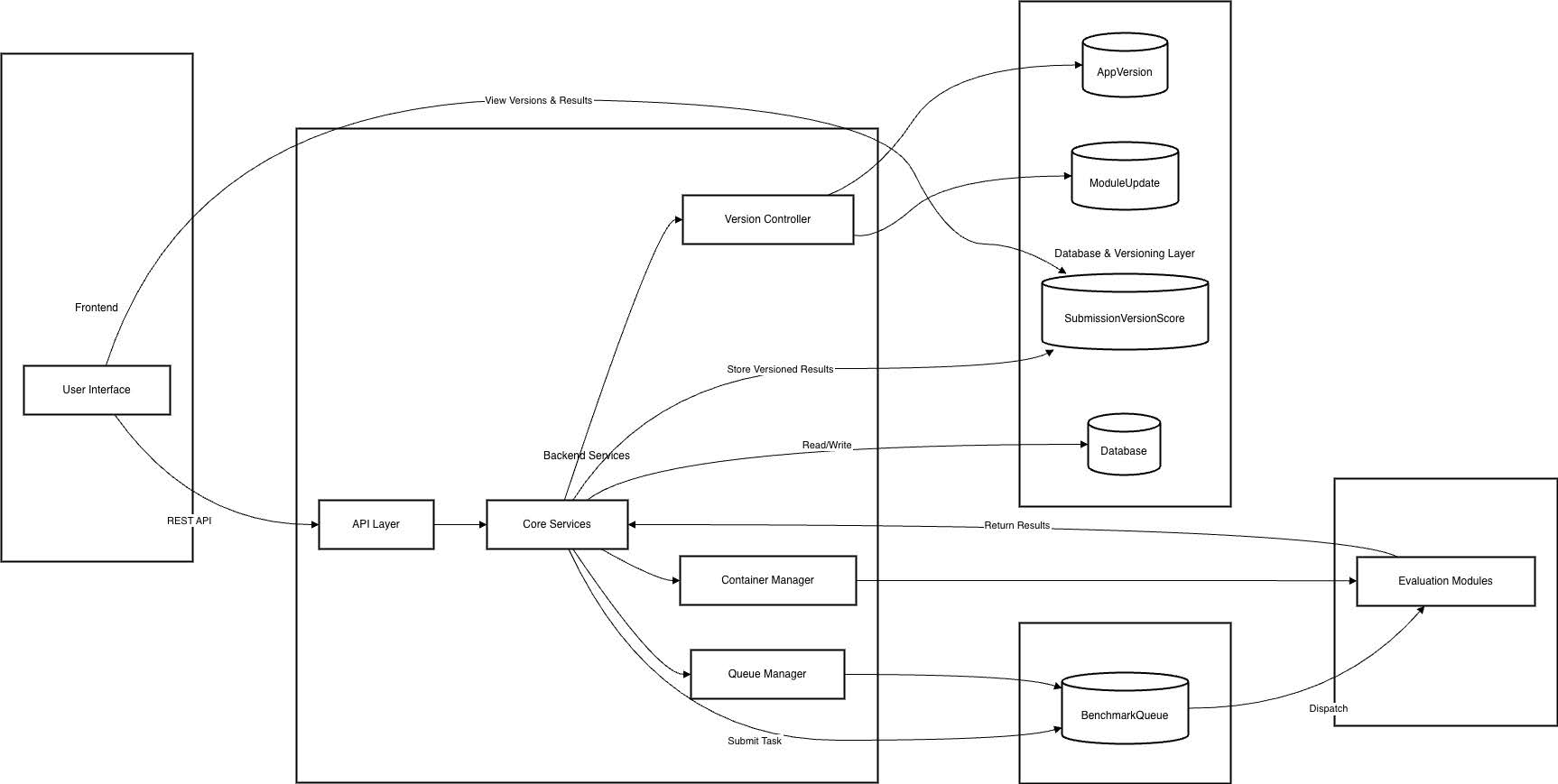}
    \caption{Component-level architecture of PrivBench.}
    \label{fig:arch}
\end{figure*}

\begin{figure*}[ht]
    \centering
    \includegraphics[width=0.95\linewidth]{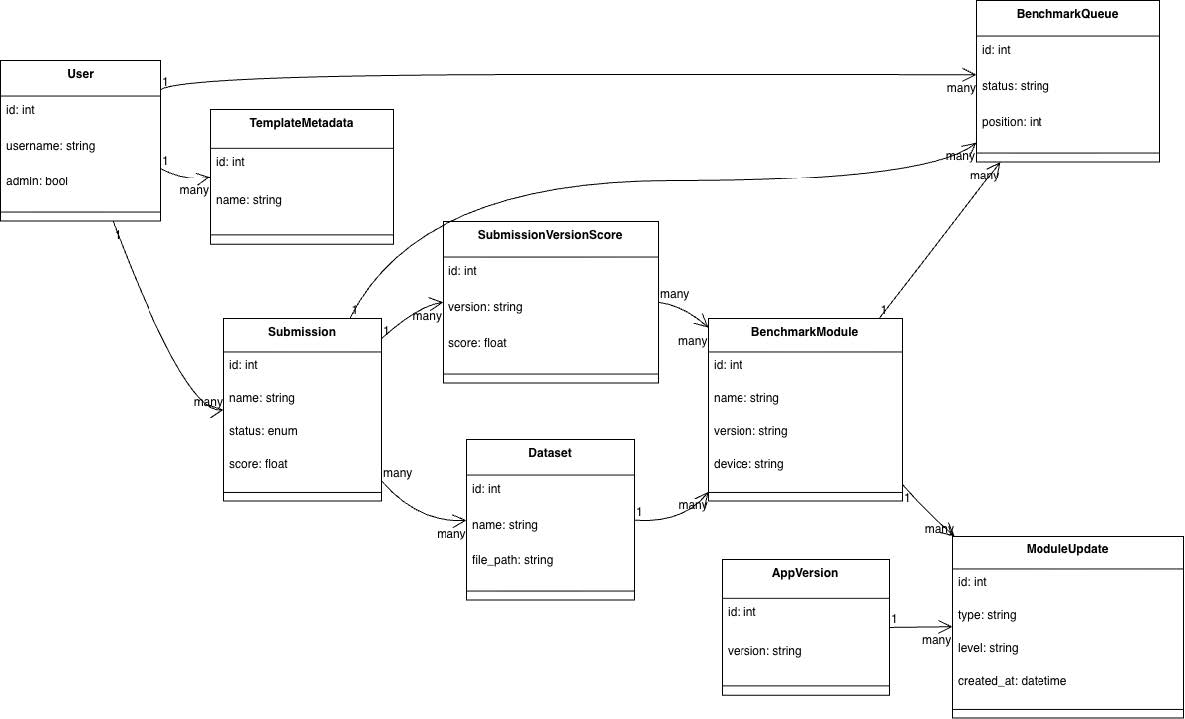}
    \caption{Simplified UML class diagram of the PrivBench web application.}
    \label{fig:class}
\end{figure*}

\end{document}